\documentclass[]{grounding-template}
\input{compile_compat.tex}
\usepackage{amsmath,amssymb}
\usepackage[shortlabels,inline]{enumitem}
\usepackage{colortbl,tabularx,pifont,wrapfig,needspace}
\setcitestyle{numbers,square,sort&compress}
\definecolor{cdsingle}{rgb}{0.941,0.953,0.910}
\definecolor{cdfull}{rgb}{0.961,0.945,0.980}
\definecolor{cdbench}{HTML}{F1F4F7}
\newcount\ArxivPreviousLines
\newcount\ArxivWrapSteps
\makeatletter
\newcommand{\FinishArxivWrap}{\par
  \ArxivWrapSteps=0
  \loop\ifnum\c@WF@wrappedlines>1
    \ArxivPreviousLines=\c@WF@wrappedlines
    \noindent\strut\par
    \advance\ArxivWrapSteps by 1
    \ifnum\c@WF@wrappedlines<\ArxivPreviousLines\else\WFclear\fi
    \ifnum\ArxivWrapSteps>80\WFclear\fi
  \repeat\WFclear}
\makeatother
\newcommand{\RewardOn}{\ding{51}}
\newcommand{\RewardOff}{\textcolor{black!35}{\ding{55}}}
\newsavebox{\ArxivInlineBox}
\newenvironment{ArxivInlineBlock}{%
  \par\FinishArxivWrap
  \begin{lrbox}{\ArxivInlineBox}\begin{minipage}{\textwidth}%
  \setlength{\parskip}{0pt}%
}{%
  \par\end{minipage}\end{lrbox}%
  \Needspace{\dimexpr\ht\ArxivInlineBox+\dp\ArxivInlineBox+\baselineskip\relax}%
  \noindent\usebox{\ArxivInlineBox}\par
}
\newtcolorbox{introquestions}{
    enhanced,
    frame hidden,
    colback=blue!5,
    borderline west={2pt}{0pt}{blue!70!black},
    sharp corners,
    boxsep=0pt,
    left=8pt,
    right=8pt,
    top=6pt,
    bottom=4pt,
    before skip=7pt plus 2pt minus 1pt,
    after skip=10pt plus 2pt minus 2pt,
}
\newcommand{\introquestion}[2]{%
    \noindent\textcolor{blue!70!black}{\textbf{\textit{Question #1:}}} #2\par\vskip3pt\relax%
}

\newcommand{\finding}[2]{
    \begin{tcolorbox}[
        colback=blue!5,
        colframe=blue!70!black,
        arc=5pt,
        boxsep=5pt,
        left=2pt,
        right=2pt,
        top=2pt,
        bottom=2pt,
        boxrule=0.8pt,
        drop shadow=gray!50!white,
        enhanced jigsaw,
        before skip=12pt plus 2pt minus 2pt, %
        after skip=12pt plus 2pt minus 2pt,  %
    ]
        \phantomsection\label{finding:#1}%
        \noindent\textbf{\textit{Finding #1:}} #2
    \end{tcolorbox}
}

\newcommand{\takeaway}[2]{
    \begin{tcolorbox}[
        colback=blue!5,
        colframe=blue!70!black,
        arc=5pt,
        boxsep=5pt,
        left=2pt,
        right=2pt,
        top=2pt,
        bottom=2pt,
        boxrule=0.8pt,
        drop shadow=gray!50!white,
        enhanced jigsaw,
        before skip=12pt plus 2pt minus 2pt, %
        after skip=12pt plus 2pt minus 2pt,  %
    ]
        \phantomsection\label{takeaway:#1}%
        \noindent\textbf{\textit{Takeaway #1:}} #2
    \end{tcolorbox}
}

\title{CodingDINO: Learning to Represent What We See as Code with Agentic Reinforcement Learning}
\begingroup
\renewcommand{\author}[2][]{%
  \ifdefempty{\authorlist}{}{\gappto\authorlist{,\hskip 0.45em\relax}}%
  \gappto\authorlist{\mbox{\authorformat[#1]{#2}}}}
\author[1,2,*]{Qize~Yu}
\author[1,*]{Hanzhen~Zhang}
\author[2,*]{Yue~Chen}
\author[1,4,*]{Jiarui~Lu}
\author[3]{Tianxing Chen}
\author[5]{Siqiao~Huang}
\author[6]{Wenxuan~Song}
\author[2]{Minghua He}
\author[7]{Yuran~Wang}
\author[2]{Junbo~Niu}
\author[1]{Zhengyu~Jia}
\author[4]{Zhihui~Shang}
\author[2]{Bowen~Ping}
\author[1]{Xianming~Liu}
\author[6]{Haoang~Li}
\author[3,\dagger]{Ping~Luo}
\author[1,\dagger]{Shuchang~Zhou}
\author[1,\dagger]{Zijian~Wang}
\endgroup
\affiliation[1]{XPeng Inc.}
\affiliation[2]{Peking University}
\affiliation[3]{The University of Hong Kong}
\affiliation[4]{Chinese Academy of Sciences}
\affiliation[5]{Tsinghua University}
\affiliation[6]{HKUST (GZ)}
\affiliation[7]{National University of Singapore}
\contribution[*]{Equal contribution}
\contribution[\dagger]{Corresponding authors}
\DeclareRobustCommand{\coverresourcelink}[1]{\href{#1}{\vphantom{#1}{\sffamily \fontsize{8.8pt}{11pt}\selectfont \texttt{#1}}}}
\project{\coverresourcelink{https://skywalker-yqz.github.io/CodingDINO/}}
\code{\coverresourcelink{https://github.com/Skywalker-yqz/CodingDINO}}
\model{\coverresourcelink{https://huggingface.co/Skywalker0410/CodingDINO}}
\titleteaser{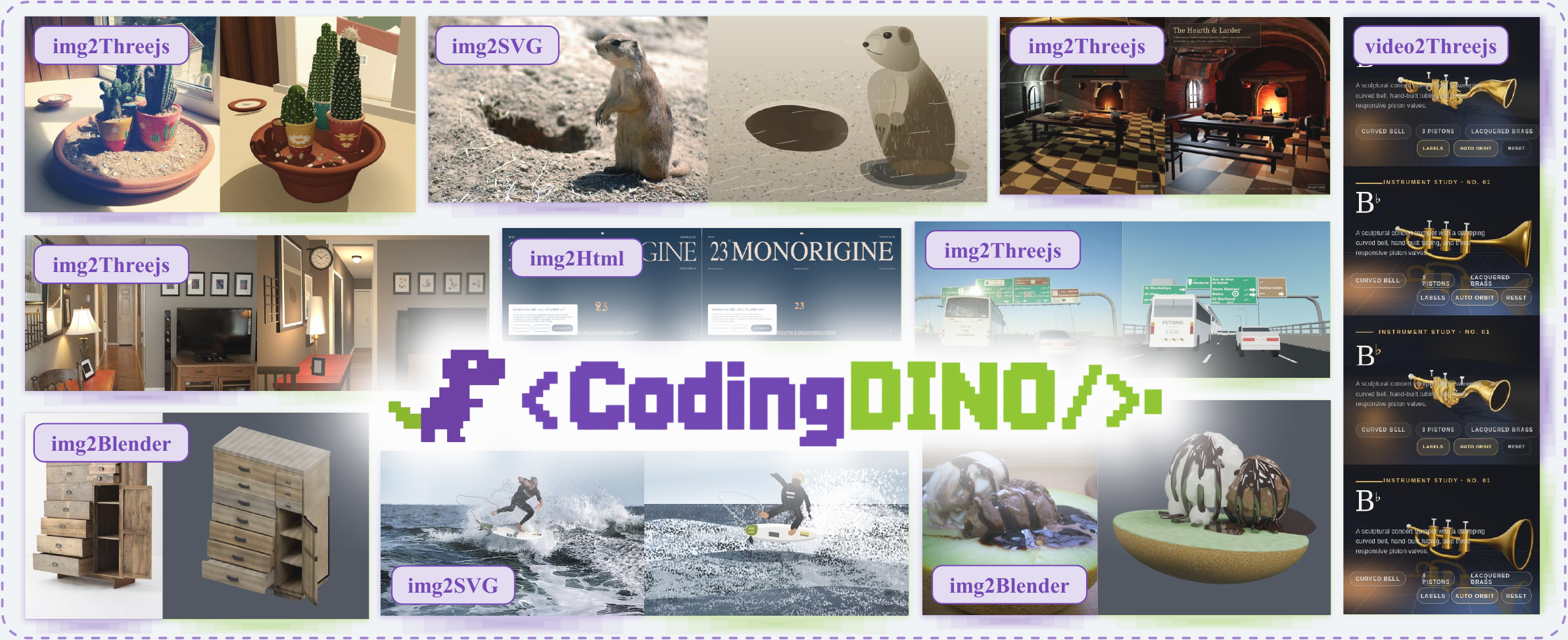}
\abstract{}

\makeatletter
\renewcommand{\mymaketitle}{%
  \tcbset{enhanced,frame hidden}%
  \tcbset{left=0.5cm}%
  \tcbset{right=0.5cm}%
  \tcbset{top=0.26cm}%
  \tcbset{bottom=0.26cm}%
  \tcbset{arc=10pt}%
  \tcbset{colback=templatebackground}%
  \tcbset{before skip=0pt}%
  \tcbset{grow to left by=1.5pt}%
  \tcbset{grow to right by=1.5pt}%
  \begin{tcolorbox}
    \hypersetup{linkcolor=templatetitle,citecolor=templatetitle,urlcolor=templatetitle}%
    \setlength{\parindent}{0cm}%
    \setlength{\parskip}{0pt}%
    {\setlength{\parskip}{0cm}%
      \raggedright
      \nohyphens
      {\setstretch{1.618}\titlelist\par}%
      \vskip 0.12cm
      \authorlist\par
      \vskip 0.12cm
      \affiliationlist\par
      \contributionlist\par
    }%
    \ifdefempty{\teaserfile}{}{%
      \smallskip
      {\setlength{\parskip}{0pt}%
        {\centering
          \includegraphics[width=\linewidth,page=1]{figures/Teaser-First.pdf}\par
        }%
        \kern-3pt
        \templatedashedrule
        \kern3pt
      }%
    }%
    \ifdefempty{\metadatalist}{}{%
      \vskip 0.04cm
      \noindent
      \begin{minipage}[c]{\linewidth}
        \setlength{\parskip}{0cm}%
        \raggedright
        \metadatalist
      \end{minipage}\par
      \vskip 0.04cm
    }%
  \end{tcolorbox}%
  \tcbset{reset}%
  \FloatBarrier
}

\makeatother

\newcommand{\codingdinotocentry}[1]{%
  \Needspace{4\baselineskip}%
  \phantomsection
  \addcontentsline{toc}{section}{#1}}
\newcommand{\listedsection}[1]{\codingdinotocentry{#1}\section*{#1}}

\begin{document}
\maketitle
\smallskip
\phantomsection
\addcontentsline{toc}{section}{Abstract}
{\centering\sffamily\bfseries Abstract\par}
\noindent Representations matter. Frontier LLMs increasingly solve complex tasks by writing and executing code, raising the question of how visual inputs can be represented in the same medium. Reconstructing images and videos as executable programs makes appearance, geometry, and temporal dynamics explicit, yielding compositional representations that LLMs can directly inspect, reason over, and edit. Existing vision-to-code methods largely emphasize single-pass generation or inference-time refinement, leaving a gap between using visual feedback and learning from it. We introduce CodingDINO, a unified agentic reinforcement learning (RL) framework for learning internal visual representations through code-based reconstruction. Inspired by DINO, our approach uses observation-to-render consistency as a general learning principle across 2D and 3D images and videos. A unified harness executes and renders SVG, HTML, ThreeJS, Blender, and animation programs, returning visual renders alongside execution diagnostics. Starting from supervised fine-tuning on verified teacher trajectories, multi-turn RL trains the agent to write, render, inspect, and revise code in this harness. Each trajectory is rewarded by execution validity and visual consistency with the target, measured using frozen DINO features and relative depth. Across six image-to-code and video-to-code benchmarks, CodingDINO rivals frontier LLMs such as Opus 5.5 and GPT-6 Astra, ranking first on 10 of 12 headline metrics and in the top three on 10. With no rendering feedback at all, its single-pass variant CodingDINO-Single retains 85.4\% of this quality on average. These results position code-based reconstruction as a route to learning reusable internal visual representations, extending DINO's representational ambition to language models that reason and act through code.
\par
\clearpage
\pdfbookmark[1]{Contents}{codingdinocontents}
\tableofcontents
\clearpage
\begin{ArxivInlineBlock}
  \captionsetup{type=figure}
  \centering
  \includegraphics[width=\linewidth]{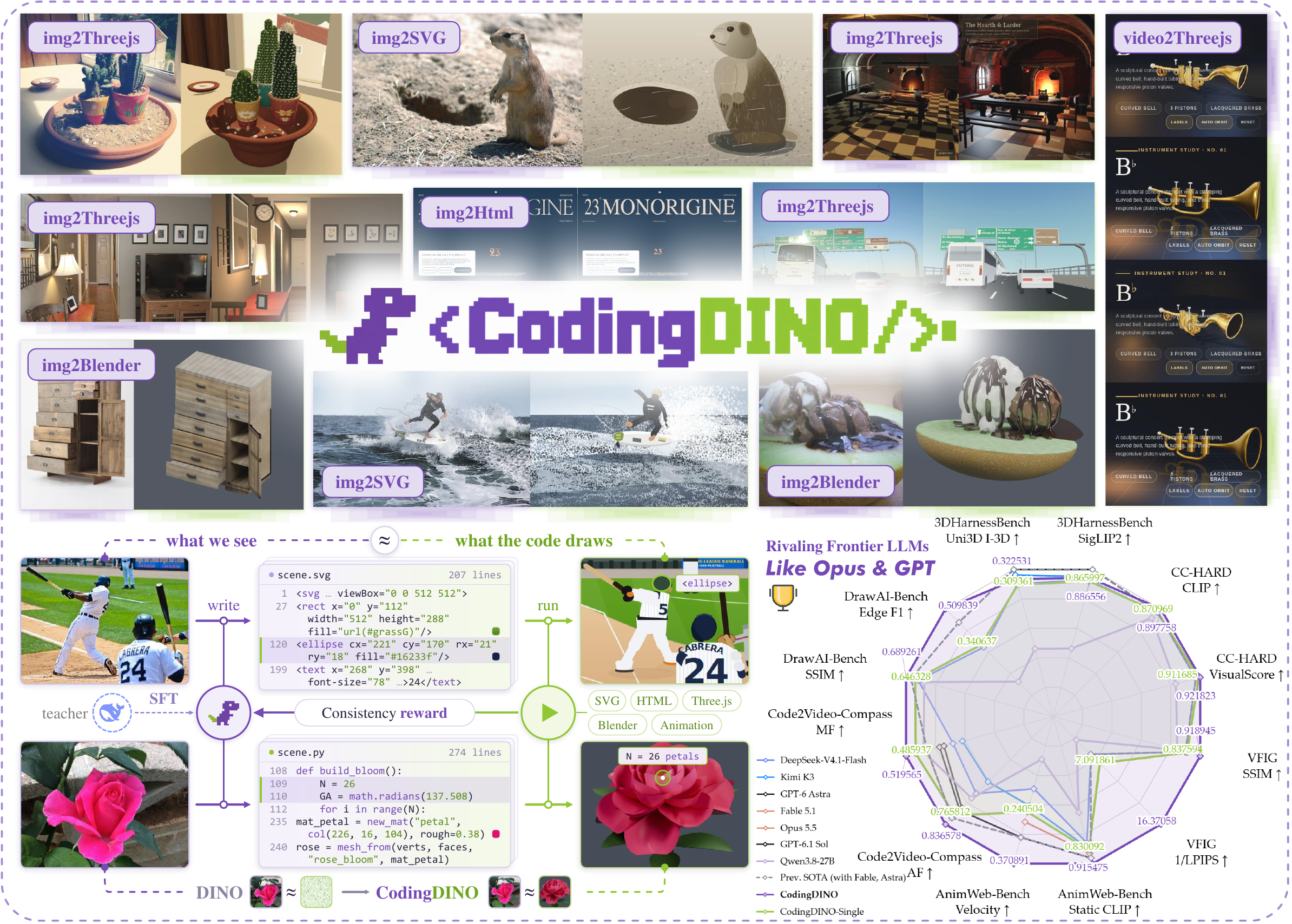}\par
  \caption{\textbf{CodingDINO represents what we see as code.} \emph{Top:} 2D/3D image and video inputs paired with program renders. \emph{Bottom:} the training consistency loop (left) and comparison with the base model and frontier LLMs (right).}\label{fig:teaser}
\end{ArxivInlineBlock}

\section{Introduction}
\label{sec:intro}

Representations matter. DINO~\cite{caron2021emergingpropertiesselfsupervisedvision} showed that a self-supervised principle, consistency between views of the same image, yields visual features that transfer across diverse tasks. These features, however, are opaque floating-point vectors, whereas frontier LLMs increasingly reason and act through code~\cite{wang2024executablecodeactionselicit,gupta2023visual,suris2023vipergpt}. A model that writes, runs, and revises programs can attend to a float vector, but it cannot execute, inspect, or edit one. We therefore ask whether what a model sees can be represented directly in the medium it manipulates best.

Code is a natural candidate. Rewriting an image or video as an executable program, a task known as vision-to-code~\cite{yang2025chartmimic,dai2026webvr,ji2026animation2code}, makes appearance, geometry, spatial relations, and motion explicit. The Blender script in Figure~\ref{fig:teaser} states that the rose has \texttt{N = 26} petals—a quantity that can be read, edited, and rendered again. Yet prior methods either refine code at inference with fixed weights~\cite{hu2024scenecraft,huang2024blenderalchemy}, distill traces into offline supervision~\cite{gao20263dcodeverse}, or apply RL only to narrow 2D domains~\cite{rodriguez2026rendering,liu2026refinesvg,li2026relook}. Two gaps remain: how to train a single vision-to-code agent online across 2D, 3D, and video domains, and whether training to inspect and revise renders only improves revision, or actually teaches the model to see better on its first draft.

DINO answers both, once in how to learn and once in what is learned. DINO trains by enforcing agreement between two views of an image; we enforce agreement between a visual observation and the render produced by its generated program. An execution-and-rendering harness makes this agreement measurable for any image or video, and agentic RL turns it into a direct training signal without requiring ground-truth source code. DINO's representation is evaluated frozen in a single forward pass. We probe ours the same way, by measuring the quality of the first program the model writes before any render comes back.

We introduce \textbf{CodingDINO}, which couples a unified visual coding harness with agentic RL (Figure~\ref{fig:method}). One harness covers 2D and 3D content, static images and dynamic videos: it executes SVG, HTML, Three.js, Blender, and animation programs and returns their renders with execution diagnostics. Training starts with supervised fine-tuning (SFT) on verified teacher programs. Multi-turn RL then lets the agent learn from its own renders, rewarding each write, render, and revise trajectory by the visual consistency between the target and its final render, measured with frozen DINO encoders and depth. The two parts need each other. The harness gives RL its observations and a label-free reward, and RL gives the harness an improvement that stays in the weights.

Across six benchmarks that span 2D and 3D, image-to-code and video-to-code, CodingDINO rivals frontier LLMs such as Opus 5.5 and GPT-6 Astra, while using 5.0--6.9$\times$ fewer tokens than the teacher DeepSeek-V4.1-Flash. It ranks first on \textbf{10 of 12} headline metrics and in the top three on 10, and on Code2Video-Compass it reaches 0.837 AF and 0.520 MF against 0.787 and 0.445 for Opus 5.5. The learning is internal. With no rendering feedback, CodingDINO-Single retains \textbf{85.4\%} of the multi-turn quality on average. Train in the loop, and the model needs the loop less: consistency alone is enough to teach a language model to represent what it sees, in a form it can run, inspect, and edit (see Appendix Figure~\ref{fig:world-model-edits}).

Our contributions are as follows.

\begin{itemize}
  \item We propose CodingDINO, a framework that jointly trains a vision-to-code model with a unified harness and agentic RL across 2D and 3D, images and videos. This self-training, driven by the consistency between what we see and what the code draws, carries DINO's label-free principle from features to programs.
  \item We study code as a visual representation for LLMs and isolate what interactive training puts into the model. As with a frozen DINO feature, we read the representation out in one pass, and we show that training in the loop improves first draft and reduces the renders needed at inference.
  \item CodingDINO surpasses its base model and its teacher and rivals frontier LLMs such as Opus 5.5 and GPT-6 Astra. It ranks first on 10 of 12 headline metrics and in the top three on 10, and its single-pass variant retains 85.4\% of this quality.
\end{itemize}

\section{Related work}
\subsection{Vision-to-code Task}

Early vision-to-code methods used domain-specific neural models and program synthesis~\cite{deng2017image,wu2017neural,ganin2018synthesizing,tian2019learning,ellis2019write} to recover UI code from screenshots (pix2code~\cite{beltramelli2018pix2code}) or graphics programs with primitives and loops (Ellis et al.~\cite{ellis2018learning}, CSGNet~\cite{geometry2022neural}), but were restricted to narrow domain-specific languages.

Recent benchmarks evaluate multimodal foundation models across webpages (Design2Code~\cite{si2025design2code}, LatCoder/CC-HARD~\cite{gui2025latcoder}), plots and editable vector graphics (Plot2Code~\cite{wu2025plot2code}, ChartMimic~\cite{yang2025chartmimic}, VFIG~\cite{he2026vfigvectorizingcomplexfigures}, DrawAI~\cite{cao2026drawaiagenticbenchmarkworkflow}), and 3D scenes (Vision2Code~\cite{periasami2026vision2code}, 3DHarnessBench~\cite{liu20263dharnessbench}). Beyond static images, IWR-Bench~\cite{chen2026iwr}, WebVR~\cite{dai2026webvr}, Animation2Code~\cite{ji2026animation2code}, AnimWeb-Bench~\cite{zhang2026animwebbenchbenchmarkinganimatedwebpage}, and Code2Video-Compass~\cite{chen2026code2video} evaluate dynamic videos, testing motion and interaction logic. Existing benchmarks treat each domain in isolation, whereas we unify 2D, 3D, static, and dynamic targets under one reconstruction formulation.

\subsection{Vision-to-code with foundation models}

A first line of work relies on offline paired supervision: WebSight~\cite{laurenccon2024unlocking} pairs synthetic webpage screenshots with HTML code, and StarVector~\cite{rodriguez2025starvector} pairs SVG programs with renderings. To exploit execution feedback, inference-time agents iteratively edit Blender code from rendered views (SceneCraft~\cite{hu2024scenecraft}, BlenderAlchemy~\cite{huang2024blenderalchemy}) without updating model weights. 3DCodeVerse~\cite{gao20263dcodeverse} builds a multi-backend 3D harness to collect verified traces for offline SFT and preference optimization, while Render-in-the-Loop~\cite{liang2026render} interleaves intermediate canvas renderings during multi-turn supervised training.

More recently, reinforcement learning has been applied to specific vision-to-code settings. RLRF~\cite{rodriguez2026rendering} optimizes single-turn SVG generation with rendering-based visual rewards, and RefineSVG~\cite{liu2026refinesvg} trains a single-step SVG revision policy using visual difference maps. ReLook~\cite{li2026relook} applies RL to front-end coding from text requests, relying on an external multimodal LLM critic for scores and textual feedback. VAGEN~\cite{wang2026vagen} studies world-model reasoning for multi-turn VLM agents with SVG reconstruction as one of several tasks.

In contrast, CodingDINO draws on DINO's visual consistency principle~\cite{caron2021emergingpropertiesselfsupervisedvision,oquab2023dinov2,simeoni2025dinov3} and code-centric visual reasoning~\cite{gupta2023visual,suris2023vipergpt,wang2024executablecodeactionselicit} to train a single agent end-to-end across 2D, 3D, and video domains. Rather than relying on offline traces, narrow 2D backends, or expensive external VLM critics, CodingDINO pairs a unified multi-domain harness with a lightweight consistency reward (frozen DINO features and relative depth), and directly isolates how interactive RL internalizes visual understanding into single-pass generation.

\section{Method}

\begin{ArxivInlineBlock}
\captionsetup{type=figure}
\centering
\includegraphics[width=0.94\textwidth]{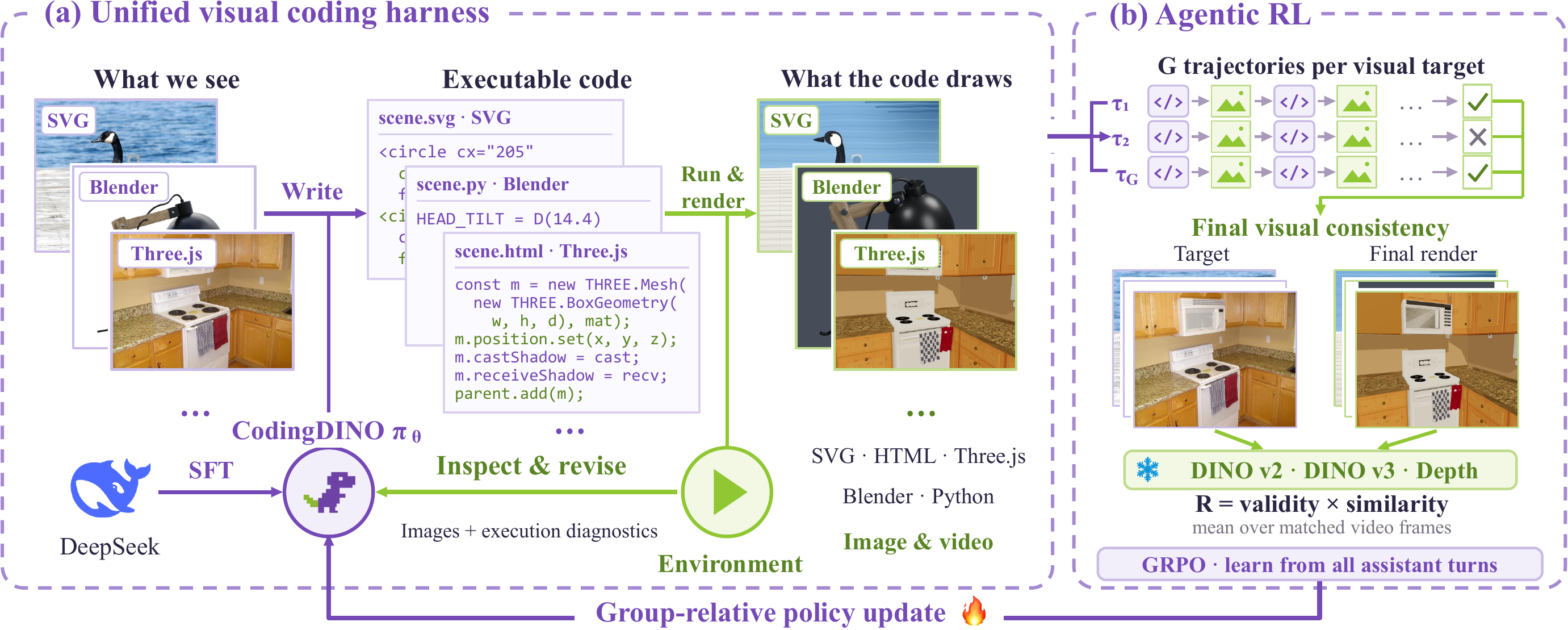}
\caption{\textbf{CodingDINO framework.} \emph{(a)} The unified harness lets the agent write, render, inspect, and revise programs using rendered images and execution diagnostics. \emph{(b)} After SFT on verified teacher trajectories, agentic RL updates the policy using visual consistency rewards between final renders and target observations.}
\label{fig:method}
\end{ArxivInlineBlock}

\subsection{Unified vision-to-code formulation}

We represent a vision-to-code task as $q=(d,u,\mathcal A,\mathcal V)$, where $d$ specifies the target program type, $u$ gives the task requirements, $\mathcal A$ contains the permitted external assets, and $\mathcal V=\{(I_i,c_i)\}_{i=1}^{M}$ is the set of target visual observations. Each $I_i$ is a target image, and $c_i$ specifies its observation conditions, including output dimensions, camera settings or viewpoint, and timestamp. A static image corresponds to a single observation, while a video corresponds to a sequence of observations ordered by time.

The output program $p$ comprises an entry file and any supporting source files. The domain-specific operator $\mathcal R_d$ executes and renders $p$ under each observation condition:

\[
\widehat I_i=\mathcal R_d(p;\mathcal A,c_i).
\]

Let $\mathcal P_d(q)$ be the set of executable programs satisfying the task constraints and producing the required observations, and let $s$ measure visual similarity. We maximize mean visual consistency across observations:

\begin{equation}
\label{eq:consistency}
\begin{gathered}
S(p,q)=\frac{1}{M}\sum_{i=1}^{M}
s\!\left(I_i,\mathcal R_d(p;\mathcal A,c_i)\right),\\
p^\star\in\underset{p\in\mathcal P_d(q)}{\arg\max}\;S(p,q).
\end{gathered}
\end{equation}

For videos, one program generates the full sequence through shared objects, state, and temporal rules.

\Needspace{4\baselineskip}
\subsection{A unified harness for visual interaction}
The harness turns any program into what the code draws and returns it to the policy, which makes the consistency in Eq.~\eqref{eq:consistency} observable during both training and inference. It maintains an isolated workspace for each task, tracks the program, the execution and rendering records, and the remaining interaction budget, and selects the execution and rendering backend by program type.

At step $t$, the policy selects action $a_t$ conditioned on the history $h_t=(q,a_1,o_1,\ldots,a_{t-1},o_{t-1})$~\citep[cf.][]{huang2026memsteer}. Given the environment state $z_t$, the harness executes $a_t$, returns observation $o_t$, and updates the state:
\[
\begin{gathered}
a_t\sim\pi_\theta(\cdot\mid h_t),\\
(z_{t+1},o_t)=\mathcal H(z_t,a_t;q).
\end{gathered}
\]
Here, $\pi_\theta$ is a multimodal language model policy with parameters $\theta$, and $\mathcal H$ is the shared harness. The agent operates through five tool groups:
\begin{itemize}
  \item \textbf{Source Inspection and Editing.} \texttt{list\_dir}, \texttt{grep\_files}, and \texttt{read\_file} locate, search, and read source files, and \texttt{apply\_patch} edits them.
  \item \textbf{Auxiliary Analysis.} \texttt{shell\_command} runs code generation and analysis scripts in isolation.
  \item \textbf{Execution and Rendering.} \texttt{render} executes and renders the program under the specified conditions and returns images or failure diagnostics.
  \item \textbf{Visual Inspection.} \texttt{view\_image} displays target and rendered images, with local crops and side-by-side comparisons of corresponding regions.
  \item \textbf{Program Submission.} \texttt{finish} submits the entry point of the verified program.
\end{itemize}
After each edit, the agent must execute and render the updated program before submission. For videos, it may inspect selected timestamps during revision, but the final rendering must cover the complete prescribed sequence.

\subsection{Agentic reinforcement learning through visual feedback}

We first perform supervised fine-tuning (SFT) on teacher trajectories whose final submissions pass execution and rendering checks, supervising program generation, revision, and tool calls at every turn. We then optimize the policy with agentic RL~\citep[cf.][]{ping2026longact,ping2026longr} on trajectories $\tau=(a_1,o_1,\ldots,a_T,o_T)$ sampled online in the harness, using final reconstruction quality.

\textbf{Visual Consistency Reward.} We measure $s$ by combining DINO feature similarity with depth consistency under matching observation conditions. Let $v(\tau,q)\in\{0,1\}$ indicate whether the trajectory explicitly submits a program whose source version has successful execution and complete rendering records. For a valid submitted program $p_\tau$, the terminal reward is

\[
R(\tau,q)=
\begin{cases}
S(p_\tau,q), & v(\tau,q)=1,\\
0, & v(\tau,q)=0.
\end{cases}
\]

\textbf{Policy Optimization.} We optimize the policy with Group Relative Policy Optimization (GRPO)~\cite{shao2024deepseekmath,yu2026groundingpi,yu2026groundanything}. For each reconstruction task $q$, the old policy $\pi_{\theta_{\mathrm{old}}}$ samples $G$ interaction trajectories in the harness. Their terminal rewards $R_i=R(\tau_i,q)$ yield the group-relative advantage $\widehat A_i=(R_i-\overline R_{-i})/(\sigma_R+\delta)$, where $\overline R_{-i}$ is the mean reward of the other trajectories for the same task, $\sigma_R$ is the group's sample standard deviation, and $\delta$ is a numerical stabilizer. Within a training batch $\mathcal B$, we apply each trajectory's advantage to its model-generated token positions $\mathcal M_i$ across all turns and maximize the core clipped objective:

\[
\begin{gathered}
J_{\mathrm{GRPO}}(\theta)=
\mathbb E_{\mathcal B}
\Biggl[
\frac{1}{Z_{\mathcal B}}
\sum_{(q,\tau_i)\in\mathcal B}
\sum_{(t,k)\in\mathcal M_i}\\[-0.4em]
\min\!\left(
r_{i,t,k}(\theta)\widehat A_i,\,
\operatorname{clip}\!\left(r_{i,t,k}(\theta),1-\epsilon,1+\epsilon\right)\widehat A_i
\right)
\Biggr].
\end{gathered}
\]

Here, $r_{i,t,k}(\theta)=\pi_\theta(a_{i,t,k}\mid h_{i,t},a_{i,t,<k})/\pi_{\theta_{\mathrm{old}}}(a_{i,t,k}\mid h_{i,t},a_{i,t,<k})$ is the probability ratio for the $k$-th generated token $a_{i,t,k}$ at turn $t$, with $a_{i,t,<k}$ denoting the preceding tokens in that turn. The normalizer $Z_{\mathcal B}=\sum_{(q,\tau_i)\in\mathcal B}|\mathcal M_i|$ counts all model-generated tokens in the batch, and $\epsilon$ specifies the clipping range.
The harness supplies visual feedback and consistency rewards, while agentic RL uses the resulting trajectories to train both initial program generation and subsequent revision.

\section{Experiments}
\label{sec:experiments}

We validate CodingDINO through experiments that answer three questions.

\begin{introquestions}
\introquestion{1}{\textbf{Performance.} Can self-training with a harness and agentic RL bring an open model to rival frontier LLMs across 2D and 3D, static and dynamic vision-to-code?}
\introquestion{2}{\textbf{Representation.} Does training in the loop improve only revision, or does it also improve the first draft through a better internal representation?}
\introquestion{3}{\textbf{Consistency.} Which visual signals make the consistency reward effective, appearance features alone or appearance together with geometry?}
\end{introquestions}

\subsection{Performance}
\label{sec:performance}

\finding{1}{Self-training with a harness and agentic RL lifts an open model past its base and its teacher to rival frontier LLMs, leading on all four headline metrics for dynamic targets. Learning inside the harness, rather than only using it at inference, is what closes the distance.}

\textbf{Setup.} We initialize CodingDINO from Qwen3.8-27B, run SFT on verified trajectories of DeepSeek-V4.1-Flash, and then train with agentic RL in the harness. We evaluate on six benchmarks that span 2D and 3D, image and video inputs: CC-HARD~\cite{gui2025latcoder} (image to HTML), VFIG-Bench~\cite{he2026vfigvectorizingcomplexfigures} (image to SVG), DrawAI-Bench~\cite{cao2026drawaiagenticbenchmarkworkflow} (image to editable graphics), 3DHarnessBench~\cite{liu20263dharnessbench} (image to Blender), AnimWeb-Bench~\cite{zhang2026animwebbenchbenchmarkinganimatedwebpage} (video to animated webpage), and Code2Video-Compass~\cite{chen2026code2video} (video to code in its video-to-video track). Following each benchmark, we use its official metrics in Tables~\ref{tab:main-3d}--\ref{tab:main-drawai}. For headline comparisons, we select one appearance metric and one structure or motion metric per benchmark and compare CodingDINO with baseline models, excluding its Single variant and training-stage ablations (see Figure~\ref{fig:radial}). \textbf{Complete results for all six benchmarks, including every reported baseline and metric, are provided in Tables~\ref{tab:main-3d}--\ref{tab:main-drawai}.} We evaluate DeepSeek-V4.1-Flash, Qwen3.8-27B, and the four CodingDINO variants ourselves; the remaining baseline results follow the cited benchmark protocols. We set thinking effort to \texttt{high} for DeepSeek-V4.1-Flash and \texttt{medium} for Qwen3.8-27B.

\textbf{CodingDINO rivals frontier LLMs on five of the six benchmarks.} Across the complete benchmark results in Tables~\ref{tab:main-3d}--\ref{tab:main-drawai}, it ranks first on 10 of the 12 headline metrics and in the top three on 10. On Code2Video-Compass, CodingDINO exceeds every reported model in both appearance and motion fidelity, including Opus 5.5 by 0.050 AF and 0.075 MF. On AnimWeb-Bench it leads GPT-6 Astra in static appearance (0.915 vs.\ 0.880 CLIP) and motion velocity (0.371 vs.\ 0.304). On static 2D targets, CodingDINO leads on CC-HARD and VFIG-Bench and ranks first on DrawAI-Bench ahead of Crafter, a multi-agent system built on GPT-5.5 (0.510 vs.\ 0.447 Edge F1).

\begin{ArxivInlineBlock}
\noindent\begin{minipage}[c]{0.32\textwidth}\raggedright
3DHarnessBench is the exception: CodingDINO scores 0.887 SigLIP2 against 0.941 for GPT-6 Astra and 0.872 for its teacher.

\textbf{Learning inside the harness adds what using it at inference cannot.} DeepSeek-V4.1-Flash and CodingDINO-Base use the same harness with fixed weights. CodingDINO surpasses Qwen3.8-27B on all 12 headline metrics, raising 3DHarnessBench SigLIP2 from 0.436 to 0.887 and DrawAI-Bench Edge F1 from 0.165 to 0.510. It surpasses its teacher on all 12, including both 3DHarnessBench metrics. In Figure~\ref{fig:qual-threejs}, CodingDINO preserves the boat's diagonal and oar arrangement. DeepSeek reverses the diagonal, and Qwen shows the boat almost broadside. See Appendix Figures~\ref{fig:qual-blender} and~\ref{fig:qual-html-svg} for Blender, HTML, and SVG examples. CodingDINO trails GPT-6 Astra on AnimWeb-Bench Static Text (0.761 vs.\ 0.780) and Qwen3.8-27B on CC-HARD TreeBLEU (0.353 vs.\ 0.421).
\end{minipage}\hfill\begin{minipage}[c]{0.65\textwidth}
\centering
\captionsetup{type=figure,justification=raggedright,singlelinecheck=false}
\includegraphics[width=\linewidth]{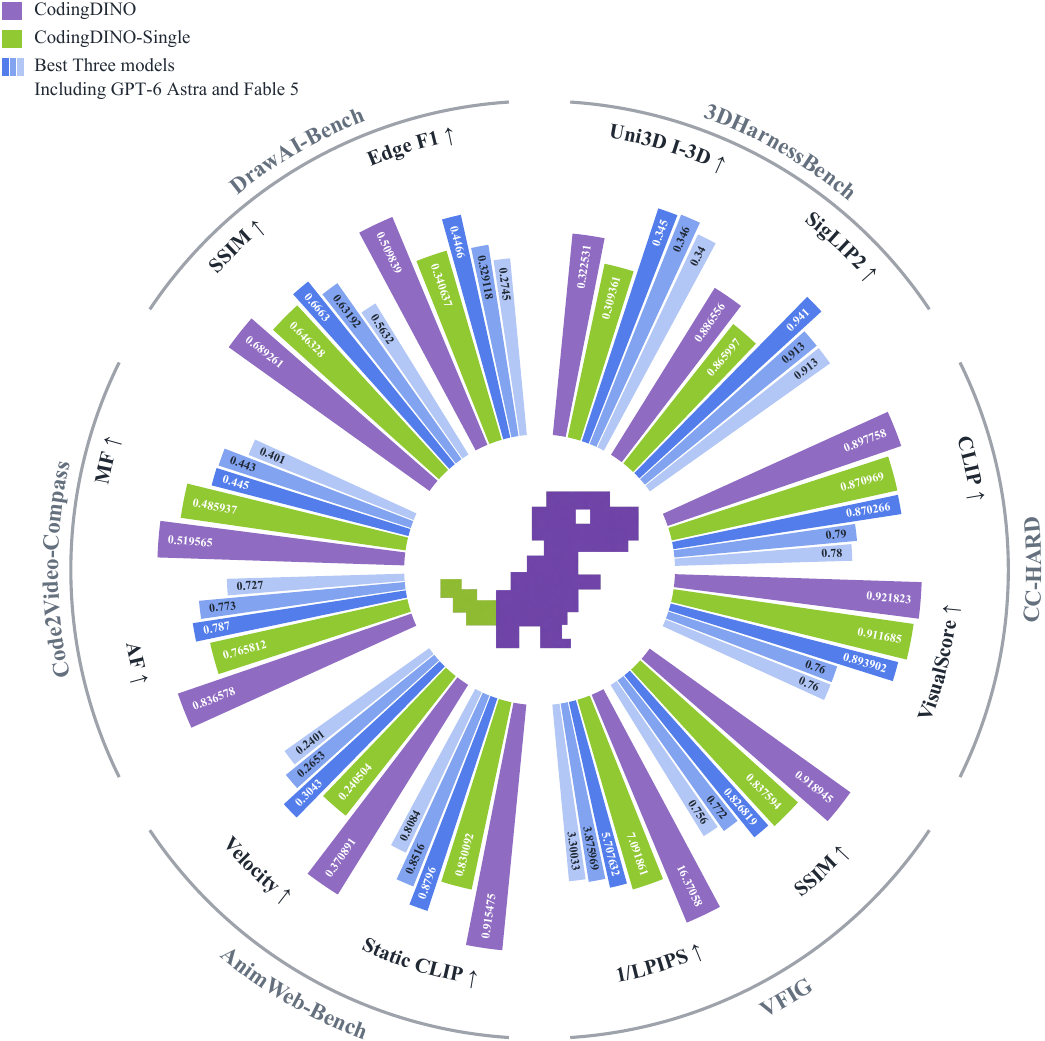}
\caption{\textbf{Reconstruction across visual coding domains.} \mbox{CodingDINO}, CodingDINO-Single, and leading baselines across six benchmarks.}
\label{fig:radial}
\end{minipage}\par\medskip
\begingroup
\captionsetup{type=table}
\centering
\caption{Complete results on 3DHarnessBench~\cite{liu20263dharnessbench}. $\dagger$ marks our evaluations; all other results are taken from the corresponding cited source. Bold marks the best value in each displayed benchmark column; -- denotes an unreported value. Headline ranks compare CodingDINO with baseline models, excluding its Single variant and training-stage ablations.}
\label{tab:main-3d}
\small
\renewcommand{\arraystretch}{1.08}
\setlength{\tabcolsep}{3pt}
\begin{tabularx}{\textwidth}{>{\raggedright\arraybackslash}Xrrrrrr}
\toprule
\rowcolor{cdbench}
\multicolumn{7}{c}{\textbf{3DHarnessBench}}\\
\midrule
Model & \shortstack{SigLIP2 $\uparrow$} & \shortstack{DINOv2-L $\uparrow$} & \shortstack{DINOv3-L $\uparrow$} & \shortstack{Uni3D\\I-3D $\uparrow$} & \shortstack{Chamfer\\official\\penalty $\downarrow$} & \shortstack{Betti\\official\\median $\downarrow$} \\
\midrule
Fable 5 & 0.913 & 0.731 & 0.784 & 0.340 & 0.012 & 0.815 \\
Opus 5 & 0.910 & 0.704 & 0.764 & 0.340 & 0.015 & 0.830 \\
GPT-5.6 Sol & 0.915 & 0.714 & 0.783 & 0.332 & 0.016 & 1.000 \\
Kimi K3 & 0.896 & 0.664 & 0.734 & 0.332 & 0.020 & 0.893 \\
Qwen 3.8 Max & 0.913 & 0.731 & 0.777 & \textbf{0.346} & 0.017 & 0.835 \\
Gemini 3.1 Pro & 0.904 & 0.692 & 0.756 & 0.335 & 0.016 & \textbf{0.776} \\
MiniMax M3 & 0.833 & 0.490 & 0.576 & 0.263 & 0.028 & 0.921 \\
GPT-6 Astra & \textbf{0.941} & \textbf{0.808} & \textbf{0.851} & 0.345 & \textbf{0.010} & 0.892 \\
\midrule
DeepSeek-V4.1-Flash$^{\dagger}$ & 0.872 & 0.615 & 0.688 & 0.312 & 0.021 & 0.935 \\
Qwen3.8-27B$^{\dagger}$ & 0.436 & 0.273 & 0.322 & 0.159 & 0.073 & 0.869 \\
CodingDINO-Base$^{\dagger}$ & 0.450 & 0.266 & 0.305 & 0.147 & 0.208 & 0.881 \\
CodingDINO-SFT$^{\dagger}$ & 0.449 & 0.244 & 0.296 & 0.146 & 0.273 & 1.000 \\
\rowcolor{cdsingle}
CodingDINO-Single$^{\dagger}$ & 0.866 & 0.621 & 0.692 & 0.309 & 0.027 & 0.851 \\
\rowcolor{cdfull}
CodingDINO$^{\dagger}$ & 0.887 & 0.648 & 0.719 & 0.323 & 0.021 & 0.971 \\
\bottomrule
\end{tabularx}
\par\endgroup

\end{ArxivInlineBlock}

\begin{table}[!tbp]
\centering
\caption{Complete results on CC-HARD~\cite{gui2025latcoder}. $\dagger$ marks our evaluations; all other results are taken from the corresponding cited source. Bold marks the best value in each displayed benchmark column; -- denotes an unreported value.}
\label{tab:main-cc-hard}
\small
\renewcommand{\arraystretch}{1.12}
\setlength{\tabcolsep}{3pt}
\begin{tabularx}{\textwidth}{>{\raggedright\arraybackslash}Xrrrrrr}
\toprule
\rowcolor{cdbench}
\multicolumn{7}{c}{\textbf{CC-HARD}}\\
\midrule
Model & \shortstack{Tree\\BLEU $\uparrow$} & \shortstack{Tree\\ROUGE $\uparrow$} & \shortstack{CLIP $\uparrow$} & \shortstack{Visual\\Score $\uparrow$} & \shortstack{Block-\\Match $\uparrow$} & \shortstack{Color $\uparrow$} \\
\midrule
DeepSeek-VL2 & 0.090 & -- & 0.750 & 0.640 & -- & -- \\
Gemini 1.5 Pro & 0.090 & -- & 0.780 & 0.760 & -- & -- \\
Gemini 1.5 Pro DCGen & 0.090 & -- & 0.730 & 0.740 & -- & -- \\
GPT-4o & 0.090 & -- & 0.790 & 0.760 & -- & -- \\
GPT-4o DCGen & 0.100 & -- & 0.740 & 0.750 & -- & -- \\
DeepSeek-VL2-tiny & 0.040 & -- & 0.660 & 0.240 & -- & -- \\
DeepSeek-VL2-small & 0.080 & -- & 0.730 & 0.610 & -- & -- \\
\midrule
DeepSeek-V4.1-Flash$^{\dagger}$ & 0.379 & 0.135 & 0.870 & 0.894 & 0.831 & 0.887 \\
Qwen3.8-27B$^{\dagger}$ & 0.421 & 0.158 & 0.822 & 0.768 & 0.566 & 0.761 \\
CodingDINO-Base$^{\dagger}$ & 0.321 & 0.119 & 0.838 & 0.771 & 0.563 & 0.784 \\
CodingDINO-SFT$^{\dagger}$ & 0.370 & 0.144 & 0.885 & 0.868 & 0.840 & 0.856 \\
\rowcolor{cdsingle}
CodingDINO-Single$^{\dagger}$ & \textbf{0.437} & \textbf{0.160} & 0.871 & 0.912 & \textbf{0.907} & 0.890 \\
\rowcolor{cdfull}
CodingDINO$^{\dagger}$ & 0.353 & 0.131 & \textbf{0.898} & \textbf{0.922} & 0.877 & \textbf{0.934} \\
\bottomrule
\end{tabularx}
\end{table}

\begin{table}[!tbp]
\centering
\caption{Complete results on VFIG-Bench~\cite{he2026vfigvectorizingcomplexfigures}. $\dagger$ marks our evaluations; all other results are taken from the corresponding cited source. Bold marks the best value in each displayed benchmark column; -- denotes an unreported value.}
\label{tab:main-vfig}
\small
\renewcommand{\arraystretch}{1.12}
\setlength{\tabcolsep}{3pt}
\begin{tabularx}{\textwidth}{>{\raggedright\arraybackslash}Xrrrrr}
\toprule
\rowcolor{cdbench}
\multicolumn{6}{c}{\textbf{VFIG-Bench}}\\
\midrule
Model & \shortstack{SSIM $\uparrow$} & \shortstack{1/LPIPS $\uparrow$} & \shortstack{VisualSim $\uparrow$} & \shortstack{Clean $\uparrow$} & \shortstack{Render $\uparrow$} \\
\midrule
Sonnet 4.6 & 0.710 & 2.123 & 0.932 & \textbf{0.856} & 0.962 \\
GPT-5.2 & 0.727 & 2.747 & 0.957 & 0.731 & 0.995 \\
Gemini 3 Flash & 0.772 & 3.876 & 0.964 & 0.788 & 0.990 \\
Gemini 3 Pro & 0.756 & 3.300 & 0.964 & 0.787 & 0.902 \\
\midrule
DeepSeek-V4.1-Flash$^{\dagger}$ & 0.827 & 5.708 & 0.971 & 0.674 & \textbf{1.000} \\
Qwen3.8-27B$^{\dagger}$ & 0.777 & 3.524 & 0.949 & 0.811 & 0.650 \\
CodingDINO-Base$^{\dagger}$ & 0.831 & 5.230 & 0.965 & 0.786 & 0.988 \\
CodingDINO-SFT$^{\dagger}$ & 0.860 & 7.334 & 0.967 & 0.718 & 0.488 \\
\rowcolor{cdsingle}
CodingDINO-Single$^{\dagger}$ & 0.838 & 7.092 & 0.975 & 0.790 & 0.998 \\
\rowcolor{cdfull}
CodingDINO$^{\dagger}$ & \textbf{0.919} & \textbf{16.371} & \textbf{0.987} & 0.701 & \textbf{1.000} \\
\bottomrule
\end{tabularx}
\end{table}

\begin{table}[!tbp]
\centering
\caption{Complete results on AnimWeb-Bench~\cite{zhang2026animwebbenchbenchmarkinganimatedwebpage}. $\dagger$ marks our evaluations; all other results are taken from the corresponding cited source. Bold marks the best value in each displayed benchmark column; -- denotes an unreported value.}
\label{tab:main-animweb}
\small
\renewcommand{\arraystretch}{1.12}
\setlength{\tabcolsep}{3pt}
\begin{tabularx}{\textwidth}{>{\raggedright\arraybackslash}Xrrrrrrrr}
\toprule
\rowcolor{cdbench}
\multicolumn{9}{c}{\textbf{AnimWeb-Bench}}\\
\midrule
Model & \shortstack{Static\\Structure $\uparrow$} & \shortstack{Static\\Text $\uparrow$} & \shortstack{Static\\Color $\uparrow$} & \shortstack{Static\\CLIP $\uparrow$} & \shortstack{Motion\\Position $\uparrow$} & \shortstack{Motion\\Size $\uparrow$} & \shortstack{Motion\\Velocity $\uparrow$} & \shortstack{Motion\\CLIP $\uparrow$} \\
\midrule
GPT-5.6 Sol & 0.560 & 0.650 & 0.660 & 0.790 & 0.358 & 0.316 & 0.231 & 0.336 \\
GPT-6 Astra & \textbf{0.697} & \textbf{0.780} & \textbf{0.745} & 0.880 & 0.453 & 0.419 & 0.304 & 0.430 \\
GPT-6 Sol & 0.575 & 0.673 & 0.656 & 0.808 & 0.373 & 0.336 & 0.240 & 0.351 \\
Fable 5.1 & 0.595 & 0.710 & 0.701 & 0.852 & 0.393 & 0.349 & 0.265 & 0.366 \\
Opus 5 & 0.518 & 0.624 & 0.646 & 0.791 & 0.312 & 0.271 & 0.205 & 0.292 \\
Grok 4.7 & 0.458 & 0.638 & 0.617 & 0.778 & 0.317 & 0.256 & 0.213 & 0.296 \\
Qwen3.8 Max & 0.427 & 0.580 & 0.559 & 0.764 & 0.302 & 0.245 & 0.203 & 0.279 \\
Gemini 3.1 Pro & 0.345 & 0.521 & 0.545 & 0.705 & 0.231 & 0.179 & 0.146 & 0.215 \\
Gemini 3.8 Flash & 0.332 & 0.506 & 0.522 & 0.679 & 0.230 & 0.176 & 0.153 & 0.212 \\
Doubao-Seed-2.1-pro & 0.338 & 0.469 & 0.515 & 0.613 & 0.206 & 0.159 & 0.124 & 0.191 \\
Kimi K3 & 0.537 & 0.677 & 0.656 & 0.820 & 0.346 & 0.298 & 0.225 & 0.322 \\
GLM-5.3-Flash & 0.429 & 0.571 & 0.575 & 0.744 & 0.295 & 0.239 & 0.196 & 0.273 \\
\midrule
DeepSeek-V4.1-Flash$^{\dagger}$ & 0.372 & 0.346 & 0.573 & 0.786 & 0.295 & 0.225 & 0.182 & 0.257 \\
Qwen3.8-27B$^{\dagger}$ & 0.287 & 0.351 & 0.464 & 0.596 & 0.171 & 0.111 & 0.107 & 0.153 \\
CodingDINO-Base$^{\dagger}$ & 0.269 & 0.262 & 0.475 & 0.662 & 0.221 & 0.159 & 0.153 & 0.191 \\
CodingDINO-SFT$^{\dagger}$ & 0.310 & 0.258 & 0.521 & 0.700 & 0.217 & 0.162 & 0.147 & 0.185 \\
\rowcolor{cdsingle}
CodingDINO-Single$^{\dagger}$ & 0.450 & 0.386 & 0.624 & 0.830 & 0.367 & 0.276 & 0.241 & 0.323 \\
\rowcolor{cdfull}
CodingDINO$^{\dagger}$ & 0.689 & 0.761 & 0.744 & \textbf{0.915} & \textbf{0.555} & \textbf{0.496} & \textbf{0.371} & \textbf{0.512} \\
\bottomrule
\end{tabularx}
\end{table}

\begin{table}[!tbp]
\centering
\caption{Complete results on Code2Video-Compass~\cite{chen2026code2video}. $\dagger$ marks our evaluations; all other results are taken from the corresponding cited source. Bold marks the best value in each displayed benchmark column; -- denotes an unreported value.}
\label{tab:main-code2video}
\small
\renewcommand{\arraystretch}{1.12}
\setlength{\tabcolsep}{3pt}
\begin{tabularx}{\textwidth}{>{\raggedright\arraybackslash}Xrrrrrrrrrr}
\toprule
\rowcolor{cdbench}
\multicolumn{11}{c}{\textbf{Code2Video-Compass}}\\
\midrule
Model & \shortstack{ES $\uparrow$} & \shortstack{TC $\downarrow$} & \shortstack{TQ $\uparrow$} & \shortstack{AQ $\uparrow$} & \shortstack{MS $\uparrow$} & \shortstack{L-TVA $\uparrow$} & \shortstack{G-TVA $\uparrow$} & \shortstack{SA $\uparrow$} & \shortstack{AF $\uparrow$} & \shortstack{MF $\uparrow$} \\
\midrule
Kimi K3 & 0.761 & 38,403 & 0.086 & 2.960 & 0.753 & 0.404 & 0.488 & 0.639 & 0.506 & 0.321 \\
MiniMax M3 & 0.686 & 11,880 & 0.068 & 2.560 & 0.680 & 0.342 & 0.418 & 0.544 & 0.410 & 0.280 \\
GPT-6 Astra & 0.850 & 12,158 & 0.107 & 3.445 & 0.840 & 0.471 & 0.557 & 0.790 & 0.727 & 0.401 \\
Opus 5.5 & 0.971 & 29,050 & 0.115 & 3.794 & 0.958 & 0.527 & \textbf{0.630} & 0.882 & 0.787 & 0.445 \\
Sonnet 5.5 & 0.963 & 34,355 & 0.118 & 3.744 & 0.948 & \textbf{0.528} & 0.555 & 0.873 & 0.773 & 0.443 \\
Haiku 5.5 & 0.931 & 38,142 & 0.107 & 3.538 & 0.921 & 0.495 & 0.580 & 0.801 & 0.668 & 0.393 \\
Gemini 3.8 Flash & 0.925 & 13,708 & 0.102 & 3.526 & 0.916 & 0.492 & 0.574 & 0.781 & 0.642 & 0.393 \\
GPT-6.1 Sol & 0.847 & 9,862 & 0.107 & 3.399 & 0.837 & 0.466 & 0.549 & 0.779 & 0.710 & 0.386 \\
GPT-6 Luna & 0.905 & \textbf{7,050} & 0.110 & 3.608 & 0.899 & 0.488 & 0.514 & 0.782 & 0.661 & 0.383 \\
Qwen3.8-Max & 0.703 & 42,217 & 0.078 & 2.685 & 0.696 & 0.376 & 0.452 & 0.616 & 0.533 & 0.317 \\
Qwen3.8-Flash & 0.709 & 26,971 & 0.079 & 2.715 & 0.702 & 0.375 & 0.454 & 0.607 & 0.504 & 0.306 \\
MiMo-V2.6-Pro & 0.879 & 36,345 & 0.108 & 3.432 & 0.870 & 0.473 & 0.558 & 0.757 & 0.624 & 0.374 \\
GLM-5.3-Flash & 0.784 & 87,794 & 0.091 & 3.067 & 0.775 & 0.424 & 0.508 & 0.684 & 0.569 & 0.345 \\
\midrule
DeepSeek-V4.1-Flash$^{\dagger}$ & 0.824 & 65,570 & 0.097 & 3.212 & 0.815 & 0.440 & 0.530 & 0.713 & 0.598 & 0.354 \\
Qwen3.8-27B$^{\dagger}$ & 0.939 & 7,114 & 0.097 & 3.471 & 0.989 & 0.513 & 0.554 & 0.792 & 0.634 & 0.459 \\
CodingDINO-Base$^{\dagger}$ & 0.636 & 10,605 & 0.093 & 3.490 & \textbf{0.990} & 0.494 & 0.496 & 0.803 & 0.642 & 0.447 \\
CodingDINO-SFT$^{\dagger}$ & 0.472 & 39,985 & 0.098 & 3.502 & 0.989 & 0.488 & 0.487 & 0.811 & 0.666 & 0.458 \\
\rowcolor{cdsingle}
CodingDINO-Single$^{\dagger}$ & \textbf{0.987} & 47,080 & 0.100 & 3.617 & 0.988 & 0.520 & 0.519 & 0.870 & 0.766 & 0.486 \\
\rowcolor{cdfull}
CodingDINO$^{\dagger}$ & 0.979 & 168,031 & \textbf{0.128} & \textbf{3.996} & 0.984 & 0.524 & 0.491 & \textbf{0.905} & \textbf{0.837} & \textbf{0.520} \\
\bottomrule
\end{tabularx}
\end{table}

\FloatBarrier
\begin{ArxivInlineBlock}
\captionsetup{type=table}
\centering
\caption{Complete results on DrawAI-Bench~\cite{cao2026drawaiagenticbenchmarkworkflow}. $\dagger$ marks our evaluations; all other results are taken from the corresponding cited source. Bold marks the best value in each displayed benchmark column; -- denotes an unreported value.}
\label{tab:main-drawai}
\small
\renewcommand{\arraystretch}{1.12}
\setlength{\tabcolsep}{3pt}
\begin{tabularx}{\textwidth}{>{\raggedright\arraybackslash}Xrrrrrrr}
\toprule
\rowcolor{cdbench}
\multicolumn{8}{c}{\textbf{DrawAI-Bench}}\\
\midrule
Model & \shortstack{SSIM $\uparrow$} & \shortstack{Edge F1 $\uparrow$} & \shortstack{MSE $\downarrow$} & \shortstack{Text F1 $\uparrow$} & \shortstack{Text box\\recall $\uparrow$} & \shortstack{Image\\region\\MSE $\downarrow$} & \shortstack{Formula\\editable $\uparrow$} \\
\midrule
Crafter (GPT-5.5) & 0.666 & 0.447 & 0.059 & 0.832 & 0.487 & 0.110 & 0.892 \\
Edit-Banana (GPT-5.5) & 0.563 & 0.275 & 0.141 & 0.159 & 0.197 & 0.227 & 0.644 \\
\midrule
DeepSeek-V4.1-Flash$^{\dagger}$ & 0.632 & 0.329 & 0.061 & 0.777 & 0.324 & 0.147 & 0.680 \\
Qwen3.8-27B$^{\dagger}$ & 0.604 & 0.165 & 0.075 & 0.760 & 0.089 & 0.178 & 0.280 \\
CodingDINO-Base$^{\dagger}$ & 0.638 & 0.200 & 0.067 & 0.783 & 0.092 & 0.172 & 0.199 \\
CodingDINO-SFT$^{\dagger}$ & 0.669 & 0.403 & 0.052 & 0.844 & 0.408 & 0.120 & 0.763 \\
\rowcolor{cdsingle}
CodingDINO-Single$^{\dagger}$ & 0.646 & 0.341 & 0.055 & \textbf{0.873} & 0.453 & 0.127 & 0.893 \\
\rowcolor{cdfull}
CodingDINO$^{\dagger}$ & \textbf{0.689} & \textbf{0.510} & \textbf{0.048} & 0.817 & \textbf{0.558} & \textbf{0.107} & \textbf{0.937} \\
\bottomrule
\end{tabularx}
\end{ArxivInlineBlock}

\subsection{Representation}
\label{sec:representation}

\finding{2}{Training in the loop improves the first draft more than it improves revision: a single pass with no rendering feedback recovers most of the multi-turn quality, and fewer renders are needed to reach it. What the loop teaches is stored in the model's representation, and the loop itself remains useful mainly for motion and fine structure.}

\textbf{The harness enables interaction; training makes its gains reusable.} CodingDINO-Base runs Qwen3.8-27B in the harness without updating its weights, separating the benefit of interaction from that of learning. It improves on Qwen3.8-27B on 10 of the 12 headline metrics, but the gains remain small on several domains: 3DHarnessBench SigLIP2 rises from 0.436 to 0.450, and DrawAI-Bench Edge F1 from 0.165 to 0.200. Full training raises these scores to 0.887 and 0.510, respectively, and improves on CodingDINO-Base on all 12 headline metrics. The harness provides observations; agentic RL teaches the model to use them.

\textbf{Supervision alone does not yield the same cross-domain behavior.} Supervised fine-tuning of Qwen3.8-27B raises SSIM from 0.777 to 0.860 on VFIG-Bench and from 0.604 to 0.669 on DrawAI-Bench. Both scores fall below those of the full model. Starting from CodingDINO-SFT, agentic RL raises the VFIG-Bench render rate from 0.488 to 1.000 and Code2Video-Compass execution success from 0.472 to 0.979. The supervised checkpoint's 3DHarnessBench SigLIP2 remains at 0.449, close to CodingDINO-Base, while AnimWeb-Bench static CLIP reaches 0.700 against 0.915 for CodingDINO. Learning from the outcomes of its own programs develops visual reconstruction capabilities that supervised imitation alone does not consistently acquire.

\FinishArxivWrap
\Needspace{18\baselineskip}
\begin{wrapfigure}{r}{0.6\textwidth}
\centering
\captionsetup{justification=raggedright,singlelinecheck=false}
\includegraphics[width=\linewidth]{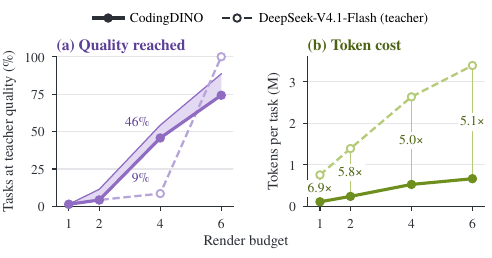}
\caption{Teacher-quality reach rate (70 selected cases; line/band: 0.01/0.03) and token cost (90 fixed cases).}
\label{fig:budget}
\end{wrapfigure}

\textbf{Most of the learned quality is already in the first draft.} CodingDINO-Single uses the fully trained model without rendering feedback and retains 85.4\% of CodingDINO's quality on average across the 12 headline metrics. It trails the multi-turn model on 3DHarnessBench and VFIG-Bench. Revision remains useful for dynamic and fine structural details: on AnimWeb-Bench, static CLIP increases from 0.830 to 0.915 and velocity from 0.241 to 0.371; on DrawAI-Bench, Edge F1 increases from 0.341 to 0.510. The improvement is therefore largely stored in the model before interaction begins.

\textbf{Strong reconstruction also requires fewer interactions.} Figure~\ref{fig:budget} shows that, within four renders, CodingDINO matches its teacher's final quality on 46\% of the selected cases (54\% with the relaxed tolerance), while the teacher has submitted on 9\%. On the fixed cost-evaluation set, CodingDINO uses 5.0--6.9 times fewer tokens across budgets of 1, 2, 4, and 6 renders. The advantage is already visible at the first render: 0.11M tokens against 0.76M for the teacher. CodingDINO commits to a useful first draft earlier and carries that advantage into subsequent refinement.

\FinishArxivWrap
\takeaway{1}{The first draft is the representation. Train in the loop, and the model needs the loop less: a program written at first sight is to a code-native LLM what a frozen feature is to DINO.}

\subsection{Consistency}
\label{sec:consistency}

\finding{3}{In the consistency reward, replacing DINOv2~\cite{oquab2023dinov2} with DINOv3~\cite{simeoni2025dinov3} or stacking the two changes little, while adding depth gives the clear gain, largest on geometry and on dynamic targets. An effective reward needs one strong appearance feature plus a geometric cue.}

We follow the setting of Section~\ref{sec:performance} and change only the encoders used in the visual consistency reward, retraining the RL stage from the same SFT checkpoint for each reward.

\textbf{The choice of DINO encoder matters little.} Across the three rewards built only from DINO features, every appearance metric in Table~\ref{tab:reward} stays within 0.02, for example 0.861 to 0.869 CLIP on CC-HARD and 0.743 to 0.759 AF on Code2Video-Compass. Stacking DINOv3 on DINOv2 does not consistently lift the structure metrics either: Edge F1 on DrawAI-Bench moves from 0.345 to 0.360, and LPIPS on VFIG-Bench stays 0.21. The one exception is Betti error on 3DHarnessBench, which falls from 28.27 with DINOv2 to 10.56 with DINOv3 and varies more than any other metric.

\begin{ArxivInlineBlock}
\captionsetup{type=table}
\centering
\caption{Visual consistency reward ablation across six benchmarks. Bold marks the best value in each column. Metric definitions follow each benchmark; LPIPS and geometry statistics retain their ablation-specific definitions.}
\label{tab:reward}
\small
\setlength{\tabcolsep}{1.8pt}
\renewcommand{\arraystretch}{1.04}
\begin{tabularx}{\textwidth}{*{3}{>{\centering\arraybackslash}X}rrrrrrr}
\toprule
\rowcolor{cdbench}
\multicolumn{10}{c}{\textbf{3DHarnessBench}}\\
\midrule
\shortstack{DINO\\v2} & \shortstack{DINO\\v3} & Depth & \shortstack{SigLIP2 $\uparrow$} & \shortstack{DINOv2-L $\uparrow$} & \shortstack{DINOv3-L $\uparrow$} & \shortstack{Uni3D $\uparrow$} & \shortstack{Uni3D\\I-3D $\uparrow$} & \shortstack{Chamfer\\mean $\downarrow$} & \shortstack{Betti L1\\mean $\downarrow$} \\
\midrule
\RewardOn & \RewardOff & \RewardOff & 0.871 & 0.598 & 0.674 & 0.593 & 0.311 & \textbf{0.021} & 28.272 \\
\RewardOff & \RewardOn & \RewardOff & 0.866 & 0.591 & 0.675 & 0.614 & \textbf{0.316} & 0.024 & 10.558 \\
\RewardOn & \RewardOn & \RewardOff & 0.879 & 0.622 & 0.695 & \textbf{0.625} & \textbf{0.316} & \textbf{0.021} & 12.930 \\
\rowcolor{cdfull}
\RewardOn & \RewardOn & \RewardOn & \textbf{0.882} & \textbf{0.625} & \textbf{0.697} & 0.622 & 0.315 & 0.023 & \textbf{7.034} \\
\bottomrule
\end{tabularx}
\par\smallskip
\begin{tabularx}{\textwidth}{*{3}{>{\centering\arraybackslash}X}rrrrrrrrrr}
\toprule
\rowcolor{cdbench}
\multicolumn{13}{c}{\textbf{CC-HARD}}\\
\midrule
\shortstack{DINO\\v2} & \shortstack{DINO\\v3} & Depth & \shortstack{Tree\\BLEU $\uparrow$} & \shortstack{Tree\\ROUGE $\uparrow$} & \shortstack{CLIP $\uparrow$} & \shortstack{Visual\\Score $\uparrow$} & \shortstack{Block-\\Match $\uparrow$} & \shortstack{Text $\uparrow$} & \shortstack{Position $\uparrow$} & \shortstack{Color $\uparrow$} & \shortstack{Visual\\CLIP $\uparrow$} & \shortstack{MAE $\downarrow$} \\
\midrule
\RewardOn & \RewardOff & \RewardOff & 0.453 & \textbf{0.174} & 0.866 & \textbf{0.911} & \textbf{0.895} & \textbf{0.975} & 0.914 & 0.894 & \textbf{0.877} & 35.068 \\
\RewardOff & \RewardOn & \RewardOff & \textbf{0.455} & 0.166 & 0.861 & 0.899 & 0.859 & 0.966 & 0.905 & 0.892 & 0.874 & 36.184 \\
\RewardOn & \RewardOn & \RewardOff & 0.433 & 0.169 & 0.869 & 0.905 & 0.866 & 0.966 & 0.915 & 0.904 & 0.876 & 32.334 \\
\rowcolor{cdfull}
\RewardOn & \RewardOn & \RewardOn & 0.354 & 0.136 & \textbf{0.880} & 0.904 & 0.861 & 0.959 & \textbf{0.917} & \textbf{0.908} & 0.875 & \textbf{24.201} \\
\bottomrule
\end{tabularx}
\par\smallskip
\begin{tabularx}{\textwidth}{*{3}{>{\centering\arraybackslash}X}rrrrrrrr}
\toprule
\rowcolor{cdbench}
\multicolumn{11}{c}{\textbf{VFIG-Bench}}\\
\midrule
\shortstack{DINO\\v2} & \shortstack{DINO\\v3} & Depth & \shortstack{SSIM $\uparrow$} & \shortstack{LPIPS $\downarrow$} & \shortstack{DINOv2-\\base $\uparrow$} & \shortstack{CLIP-B/32 $\uparrow$} & \shortstack{SigLIP2 $\uparrow$} & \shortstack{VisualSim $\uparrow$} & \shortstack{Clean $\uparrow$} & \shortstack{Render $\uparrow$} \\
\midrule
\RewardOn & \RewardOff & \RewardOff & 0.770 & 0.210 & 0.956 & 0.949 & 0.982 & 0.963 & \textbf{0.789} & \textbf{1.000} \\
\RewardOff & \RewardOn & \RewardOff & 0.772 & 0.214 & 0.949 & 0.945 & 0.979 & 0.957 & \textbf{0.789} & 0.983 \\
\RewardOn & \RewardOn & \RewardOff & 0.762 & 0.213 & 0.955 & 0.951 & 0.982 & 0.962 & 0.766 & 0.998 \\
\rowcolor{cdfull}
\RewardOn & \RewardOn & \RewardOn & \textbf{0.802} & \textbf{0.162} & \textbf{0.965} & \textbf{0.960} & \textbf{0.987} & \textbf{0.971} & 0.782 & 0.995 \\
\bottomrule
\end{tabularx}
\par\smallskip
\begin{tabularx}{\textwidth}{*{3}{>{\centering\arraybackslash}X}rrrrrrrr}
\toprule
\rowcolor{cdbench}
\multicolumn{11}{c}{\textbf{AnimWeb-Bench}}\\
\midrule
\shortstack{DINO\\v2} & \shortstack{DINO\\v3} & Depth & \shortstack{Static\\Structure $\uparrow$} & \shortstack{Static\\Text $\uparrow$} & \shortstack{Static\\Color $\uparrow$} & \shortstack{Static\\CLIP $\uparrow$} & \shortstack{Motion\\Position $\uparrow$} & \shortstack{Motion\\Size $\uparrow$} & \shortstack{Motion\\Velocity $\uparrow$} & \shortstack{Motion\\CLIP $\uparrow$} \\
\midrule
\RewardOn & \RewardOff & \RewardOff & 0.425 & 0.404 & 0.599 & 0.804 & 0.361 & 0.264 & 0.232 & 0.321 \\
\RewardOff & \RewardOn & \RewardOff & 0.427 & \textbf{0.438} & 0.595 & 0.795 & 0.360 & 0.261 & 0.230 & 0.319 \\
\RewardOn & \RewardOn & \RewardOff & 0.435 & 0.413 & 0.597 & 0.814 & 0.370 & 0.271 & 0.232 & 0.324 \\
\rowcolor{cdfull}
\RewardOn & \RewardOn & \RewardOn & \textbf{0.484} & 0.393 & \textbf{0.672} & \textbf{0.884} & \textbf{0.393} & \textbf{0.311} & \textbf{0.260} & \textbf{0.345} \\
\bottomrule
\end{tabularx}
\par\smallskip
\begin{tabularx}{\textwidth}{*{3}{>{\centering\arraybackslash}X}rrrrrrrrrr}
\toprule
\rowcolor{cdbench}
\multicolumn{13}{c}{\textbf{Code2Video-Compass}}\\
\midrule
\shortstack{DINO\\v2} & \shortstack{DINO\\v3} & Depth & \shortstack{ES $\uparrow$} & \shortstack{TC $\downarrow$} & \shortstack{TQ $\uparrow$} & \shortstack{AQ $\uparrow$} & \shortstack{MS $\uparrow$} & \shortstack{L-TVA $\uparrow$} & \shortstack{G-TVA $\uparrow$} & \shortstack{SA $\uparrow$} & \shortstack{AF $\uparrow$} & \shortstack{MF $\uparrow$} \\
\midrule
\RewardOn & \RewardOff & \RewardOff & \textbf{0.998} & 24,644 & 0.101 & 3.619 & \textbf{0.989} & 0.513 & 0.516 & 0.857 & 0.746 & 0.481 \\
\RewardOff & \RewardOn & \RewardOff & 0.965 & \textbf{19,117} & 0.101 & 3.561 & \textbf{0.989} & 0.513 & 0.514 & 0.855 & 0.743 & 0.472 \\
\RewardOn & \RewardOn & \RewardOff & 0.973 & 38,250 & 0.101 & \textbf{3.624} & 0.987 & 0.521 & 0.519 & 0.861 & 0.759 & 0.480 \\
\rowcolor{cdfull}
\RewardOn & \RewardOn & \RewardOn & 0.970 & 72,429 & \textbf{0.102} & 3.612 & 0.985 & \textbf{0.525} & \textbf{0.520} & \textbf{0.885} & \textbf{0.803} & \textbf{0.496} \\
\bottomrule
\end{tabularx}
\par\smallskip
\begin{tabularx}{\textwidth}{*{3}{>{\centering\arraybackslash}X}rrrrrrrrr}
\toprule
\rowcolor{cdbench}
\multicolumn{12}{c}{\textbf{DrawAI-Bench}}\\
\midrule
\shortstack{DINO\\v2} & \shortstack{DINO\\v3} & Depth & \shortstack{SSIM $\uparrow$} & \shortstack{Edge F1 $\uparrow$} & \shortstack{MSE $\downarrow$} & \shortstack{Text F1 $\uparrow$} & \shortstack{Text box\\recall $\uparrow$} & \shortstack{Image\\source\\recall $\uparrow$} & \shortstack{Image IoU $\uparrow$} & \shortstack{Image\\region\\MSE $\downarrow$} & \shortstack{Formula\\editable $\uparrow$} \\
\midrule
\RewardOn & \RewardOff & \RewardOff & \textbf{0.657} & 0.345 & 0.054 & \textbf{0.869} & 0.398 & 0.028 & \textbf{0.000} & 0.133 & 0.859 \\
\RewardOff & \RewardOn & \RewardOff & 0.638 & 0.343 & 0.057 & 0.860 & 0.384 & 0.029 & \textbf{0.000} & 0.140 & 0.837 \\
\RewardOn & \RewardOn & \RewardOff & 0.639 & 0.360 & 0.058 & 0.861 & 0.416 & 0.023 & \textbf{0.000} & 0.134 & 0.723 \\
\rowcolor{cdfull}
\RewardOn & \RewardOn & \RewardOn & 0.654 & \textbf{0.394} & \textbf{0.052} & 0.866 & \textbf{0.483} & \textbf{0.030} & \textbf{0.000} & \textbf{0.123} & \textbf{0.873} \\
\bottomrule
\end{tabularx}
\end{ArxivInlineBlock}

\FinishArxivWrap
\Needspace{17\baselineskip}
\begin{wrapfigure}{r}{0.6\textwidth}
\centering
\captionsetup{justification=raggedright,singlelinecheck=false}
\includegraphics[width=\linewidth]{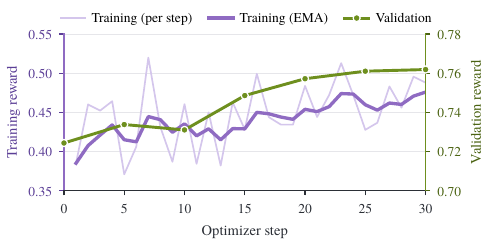}
\caption{Training reward (raw and EMA, left) and validation reward (right) during agentic RL.}
\label{fig:reward}
\end{wrapfigure}

\textbf{Depth brings the gain, mostly in structure and motion.} Adding depth to the two DINO encoders lowers pixel MAE on CC-HARD from 32.33 to 24.20, LPIPS on VFIG-Bench from 0.213 to 0.162, and the Betti error on 3DHarnessBench from 12.93 to 7.03, and it raises Edge F1 on DrawAI-Bench from 0.360 to 0.394. On the two video benchmarks, depth raises static CLIP from 0.814 to 0.884 and velocity from 0.232 to 0.260 on AnimWeb-Bench, and AF from 0.759 to 0.803 on Code2Video-Compass. Embedding-level appearance on static images changes less, from 0.879 to 0.882 SigLIP2 on 3DHarnessBench and from 0.982 to 0.987 on VFIG-Bench. Chamfer distance on 3DHarnessBench stays between 0.021 and 0.024 across all four rewards, and TreeBLEU on CC-HARD drops from 0.433 to 0.354. We hypothesize that depth helps because it separates overlapping layers and moving objects that global DINO features summarize together.

Both rewards rise over the first 30 optimizer steps (Figure~\ref{fig:reward}), with validation reward increasing from 0.724 to 0.762 while the share of valid submissions reaches 100\%.

\FinishArxivWrap
\takeaway{2}{Consistency is supervision enough. DINO learns from the agreement between two views of an image, and CodingDINO learns from the agreement between what we see and what the code draws. A harness with agentic RL turns that agreement into self-training across 2D and 3D, images and videos.}

\FinishArxivWrap
\FloatBarrier
\section{Conclusion}

We presented CodingDINO, a unified agentic reinforcement learning framework for reconstructing images and videos as executable 2D and 3D programs. By learning online from execution diagnostics and visual consistency between target observations and renders—combining frozen DINO features with relative depth for geometry and motion—CodingDINO rivals frontier LLMs across six benchmarks. Its single-pass variant retains 85.4\% of multi-turn quality on average without rendering feedback, while interactive inference cuts token cost by 5.0--6.9$\times$ over the teacher, showing that interactive training improves the model's first draft. These findings support code-based reconstruction as a route to learning reusable visual representations: training in the loop reduces reliance on the loop at inference.

\listedsection{LLM Usage Statement}
We used LLMs, including OpenAI (ChatGPT / GPT), Anthropic (Claude), Google (Gemini), to assist with manuscript formatting, language polishing, code implementation and debugging, experimental data organization and visualization, and preparation of implementation details and hyperparameter descriptions in the supplementary material. Additionally, LLM-based teacher models were employed for training data and trajectory generation, as detailed in our methodology. All research decisions, experimental results, technical claims, and final manuscript content were reviewed and verified by the authors.

\listedsection{Ethics Statement}
This work studies the reconstruction of visual content as executable programs. Generated programs should be executed in isolated environments, with external assets and permissions reviewed before use. Images, videos, datasets, and model outputs should be used in accordance with applicable licenses, privacy protections, and intellectual-property rights. We encourage attribution of source material and clear disclosure of synthetic or reconstructed content.

\listedsection{Reproducibility Statement}
Appendix~\ref{sec:supp-metrics} defines the benchmark metrics and the calculation of single-pass quality retention. Appendix~\ref{sec:supp-training} gives the reconstruction prompts, SFT and multi-turn GRPO settings, appearance and depth reward definitions, and evaluation protocols for render budgets and token use.

\listedsection{Research Scope, Data Use, and Institutional Disclaimer}
We conducted this work solely for scientific investigation and academic publication. We used the data to study how models learn visual representations through code-based reconstruction. Commercial product development and deployment fall outside the scope of this research. The original licenses and terms of use continue to govern all third-party datasets and model resources.

The views, methods, findings, and conclusions presented in this paper are those of the authors and do not represent the official positions, product plans, or commercial commitments of XPeng Inc. or the other affiliated institutions. Institutional affiliations do not imply endorsement of downstream applications.

\listedsection{Acknowledgments}
We thank Xpeng Inc. for providing computational and data resources in support of this academic research, and the data team for their assistance with data preparation and research support. We are particularly grateful to Professor Ping Luo for his guidance on the research ideas and manuscript writing. We also thank Hang Zhang, Weiyang Jin, Mengfan Dong, Xirui Liang, Tianshu Wu, Jiadi You, Linhao Zhou, Qiman Wu, Ziteng Cui, Haojun Zhang, Min Chen, Hao Li, and Yi Zou for their valuable suggestions and constructive feedback.

\FinishArxivWrap
\FloatBarrier
\preto\bibsection{\codingdinotocentry{\refname}}
\bibliographystyle{ieeenat_fullname}
\bibliography{main}
\clearpage
\beginappendix
\addtocontents{toc}{\protect\setcounter{tocdepth}{1}}
\section{Complete experimental results}
\label{sec:supp-results}

\Needspace{4\baselineskip}
\subsection{Benchmarks and metrics}
\label{sec:supp-metrics}
Tables~\ref{tab:main-3d}--\ref{tab:main-drawai} report all six benchmarks. A dagger marks models we evaluated. Other rows follow the benchmark cited in each caption. Arrows indicate the preferred direction, bold marks the best displayed value including ties, and a dash denotes an unreported result.

\paragraph{Reading the headline metrics.}
The headline pairs are SigLIP2 and Uni3D I-3D on 3DHarnessBench, CLIP and VisualScore on CC-HARD, SSIM and reciprocal LPIPS on VFIG-Bench, static CLIP and velocity on AnimWeb-Bench, AF and MF on Code2Video-Compass, and SSIM and Edge F1 on DrawAI-Bench. We rank CodingDINO against the baseline models and use the other CodingDINO variants for within-model comparisons. Let $s_j$ and $m_j$ denote the single-pass and multi-turn scores on headline metric $j$, respectively. All twelve scores increase with quality. The mean fraction of multi-turn quality retained in the first draft is
\begin{equation}
  \frac{1}{12}\sum_{j=1}^{12}\frac{s_j}{m_j}\simeq 0.854.
\end{equation}
We compute the ratios before rounding the benchmark scores. Each metric contributes equally, independent of its numerical scale.

\paragraph{From the base model to the teacher and the trained agent.}
Across the six benchmarks, CodingDINO surpasses Qwen3.8-27B on all 12 headline metrics and its trajectory teacher, DeepSeek-V4.1-Flash, on all 12. On DrawAI-Bench, Edge F1 rises from 0.165 for Qwen to 0.329 for DeepSeek and 0.510 for CodingDINO, indicating closer agreement with the reference contours. On AnimWeb-Bench, the same three models score 0.107, 0.182 and 0.371 on velocity: the motion in the trained agent's animations more closely matches the target velocity. The supervised checkpoint scores 0.147 on velocity. After agentic RL, the first draft reaches 0.241, overtaking the teacher before the agent sees a render. CodingDINO leads the teacher on both 3DHarnessBench headline metrics, as it does on each of the other five benchmarks (Tables~\ref{tab:main-3d}--\ref{tab:main-drawai}).

\paragraph{3DHarnessBench.}
SigLIP2, DINOv2-L, and DINOv3-L measure appearance by averaging image-feature cosine similarity across four matched views. Uni3D compares the target and reconstructed point clouds, while Uni3D I-3D compares the target image with the reconstructed point cloud. Chamfer measures the squared distance from each surface to its nearest points on the other after alignment. A missing mesh incurs $1.5$ times the largest valid Chamfer distance in the run. Betti measures differences in the number of connected components, loops, and enclosed cavities. The official score is the median of $\|\hat{\beta}-\beta\|_1/\|\beta\|_1$, where $\beta$ and $\hat{\beta}$ are the target and reconstructed Betti vectors. In a single pass, CodingDINO-Single reconstructs both appearance and surface geometry more accurately than CodingDINO-SFT. DINOv3-L similarity rises from 0.296 to 0.692 and penalized Chamfer distance falls from 0.273 to 0.027 (Table~\ref{tab:main-3d}).

\paragraph{CC-HARD.}
TreeBLEU measures the fraction of generated parent--children tag patterns that occur in the reference HTML tree. TreeROUGE measures the fraction of distinct reference patterns recovered. CLIP compares whole-page screenshot embeddings, while mean absolute error (MAE) measures pixel color differences. VisualScore averages five components. Block-Match measures the share of text-block area covered by matched blocks across the two pages. Text, Position, and Color compare the content, centers, and lettering color of those blocks. Visual CLIP compares the screenshots with text regions masked. CodingDINO-Single matches 90.7\% of the text-block area on average in its first draft, up from 56.3\% for CodingDINO-Base. VisualScore rises from 0.771 to 0.912. CodingDINO also attains the highest Color score, 0.934 (Table~\ref{tab:main-cc-hard}).

\paragraph{VFIG-Bench.}
SSIM measures local luminance, contrast and structure in a rasterized SVG. Reciprocal LPIPS increases as the perceptual distance between the rendering and the reference falls. VisualSim averages their cosine similarities under the DINOv2-base, CLIP-B/32 and SigLIP2 image encoders. The Render score is the fraction of SVGs that render successfully. For a figure with $N$ geometric elements, $B$ basic shapes and $K$ connectors, $\mathrm{Clean}=(B+K)/N$ measures how much of its geometry is expressed with editable primitives. In Table~\ref{tab:main-vfig}, CodingDINO-Single renders 99.8\% of its first drafts and reaches 0.975 VisualSim, close to the full policy's 0.987. Its first draft is perceptually closer to the reference than the teacher's output (reciprocal LPIPS, 7.092 vs.\ 5.708) and represents a larger share of the figure with editable primitives (Clean, 0.790 vs.\ 0.674).

\paragraph{AnimWeb-Bench.}
On aligned keyframes, Structure compares the overlap, centers, areas and aspect ratios of corresponding interface elements. Text measures edit similarity between recognized strings in reading order, Color compares foreground and background colors, and Static CLIP compares the whole frames. Between keyframes, the benchmark tracks individual elements. Position scores their center locations, Size compares their widths and heights, Velocity compares their direction and speed, and Motion CLIP compares their appearance within the tracked boxes. Unmatched elements lower the trajectory scores. From the first draft to the revised animation, CodingDINO improves all four trajectory scores (Table~\ref{tab:main-animweb}). Revision brings element scale and velocity closer to the target, raising Size from 0.276 to 0.496 and Velocity from 0.241 to 0.371.

\paragraph{Code2Video-Compass.}
In the video-to-video track, execution success (ES) is the fraction of programs that produce a playable video, and token consumption (TC) is the mean output token count per attempted task. Technical quality (TQ) scores clarity, aesthetic quality (AQ) scores frame aesthetics, and motion smoothness (MS) measures how accurately adjacent frames predict the intermediate frame. Local and global text--video alignment (L-TVA and G-TVA) measure agreement with instruction segments and with the full instruction, respectively. Three scores compare the output directly with the reference clip. Sequence alignment (SA) averages DINOv3 cosine similarity along the temporally aligned frame sequence. Appearance fidelity (AF) measures perceptual agreement along a DreamSim alignment, and motion fidelity (MF) compares the directions and magnitudes of tracked point displacements. CodingDINO leads the reported models on all three measures, with 0.905 SA, 0.837 AF and 0.520 MF (Table~\ref{tab:main-code2video}). Its first draft already reaches 0.486 MF, exceeding the 0.445 of Opus~5.5 before any rendering feedback.

\paragraph{DrawAI-Bench.}
SSIM compares local luminance, contrast, and structure. Pixel MSE measures mean squared color error over the entire canvas, while image-region MSE averages that error within annotated image boxes. Edge F1 and Text F1 combine precision and recall as $2PR/(P+R)$. Edge matches allow one pixel of displacement, and text matches compare normalized strings. Text-box recall counts correctly localized editable text boxes relative to the reference, and formula editability measures the fraction of reference formulas recovered in editable form. Image source recall measures recovery of annotated image assets, and image IoU measures overlap between matched image boxes, counting missing boxes as zero. CodingDINO-Single attains 0.873 Text F1 and 0.893 formula editability. With revision, text-box recall increases from 0.453 to 0.558 and Edge F1 from 0.341 to 0.510. The first draft preserves editable content, and visual feedback improves placement and contours (Table~\ref{tab:main-drawai}).

\Needspace{4\baselineskip}
\subsection{Harness and training-stage comparisons}
\label{sec:supp-stage-comparisons}

\paragraph{Harness gains with fixed weights.}
CodingDINO-Base runs Qwen3.8-27B in the interactive harness with its weights fixed. This comparison measures what execution and visual feedback contribute at inference. The harness improves 10 of the 12 headline metrics (Table~\ref{tab:supp-additional-checkpoint-metrics}). On VFIG-Bench, the fraction of programs that render rises from 0.650 to 0.988, while SSIM rises from 0.777 to 0.831. The agent can repair an SVG and bring its appearance closer to the input within the same episode.

\paragraph{Total training gain relative to Qwen3.8-27B.}
The fully trained CodingDINO agent improves on the starting model on all 12 headline metrics. On 3DHarnessBench, SigLIP2 rises from 0.436 to 0.887 and image-to-3D similarity from 0.159 to 0.323. With the same harness, CodingDINO-Base scores 0.450 and 0.147 on these measures. Most of the gain in matching the object's appearance across viewpoints and recovering its 3D shape comes after training.

\paragraph{SFT gains relative to Qwen3.8-27B.}
The checkpoint trained on verified teacher trajectories exceeds Qwen3.8-27B on 10 of the 12 headline metrics. Its clearest gains are on static graphics. VFIG-Bench reciprocal LPIPS rises from 3.524 to 7.334, and DrawAI-Bench Edge F1 rises from 0.165 to 0.403. These gains show that supervised trajectories teach the policy to reproduce perceptual detail and place contours accurately.

\begin{ArxivInlineBlock}
\captionsetup{type=table}
\centering\small
\caption{Reconstruction across five model settings. Qwen3.8-27B is the starting policy. CodingDINO-Base adds the harness, CodingDINO-SFT adds supervised training, and CodingDINO-Single and CodingDINO use the same RL-trained weights with single-pass and interactive inference, respectively. Bold marks the best reported value in each row. A dash indicates an unreported result on the same split.}
\label{tab:supp-additional-checkpoint-metrics}
\setlength{\tabcolsep}{3pt}
\begin{tabular*}{\textwidth}{@{\extracolsep{\fill}}llrrr>{\columncolor{cdsingle}}r>{\columncolor{cdfull}}r@{}}
\toprule
\rowcolor{cdbench}
Benchmark & Metric & \shortstack{Qwen3.8-\\27B} & \shortstack{CodingDINO-\\Base} & \shortstack{CodingDINO-\\SFT} & \shortstack{CodingDINO-\\Single} & CodingDINO \\
\midrule
\multicolumn{7}{l}{\emph{Headline metrics and program execution}} \\
3DHarnessBench & SigLIP2 $\uparrow$ & 0.436 & 0.450 & 0.449 & 0.866 & \textbf{0.887} \\
 & Uni3D I-3D $\uparrow$ & 0.159 & 0.147 & 0.146 & 0.309 & \textbf{0.323} \\
\midrule
CC-HARD & CLIP $\uparrow$ & 0.822 & 0.838 & 0.885 & 0.871 & \textbf{0.898} \\
 & VisualScore $\uparrow$ & 0.768 & 0.771 & 0.868 & 0.912 & \textbf{0.922} \\
\midrule
VFIG-Bench (ID) & SSIM $\uparrow$ & 0.777 & 0.831 & 0.860 & 0.838 & \textbf{0.919} \\
 & 1/LPIPS $\uparrow$ & 3.524 & 5.230 & 7.334 & 7.092 & \textbf{16.371} \\
 & Render $\uparrow$ & 0.650 & 0.988 & 0.488 & 0.998 & \textbf{1.000} \\
\midrule
AnimWeb-Bench & Static CLIP $\uparrow$ & 0.596 & 0.662 & 0.700 & 0.830 & \textbf{0.915} \\
 & Velocity $\uparrow$ & 0.107 & 0.153 & 0.147 & 0.241 & \textbf{0.371} \\
\midrule
Code2Video-Compass & AF $\uparrow$ & 0.634 & 0.642 & 0.666 & 0.766 & \textbf{0.837} \\
 & MF $\uparrow$ & 0.459 & 0.447 & 0.458 & 0.486 & \textbf{0.520} \\
 & ES $\uparrow$ & 0.939 & 0.636 & 0.472 & \textbf{0.987} & 0.979 \\
\midrule
DrawAI-Bench & SSIM $\uparrow$ & 0.604 & 0.638 & 0.669 & 0.646 & \textbf{0.689} \\
 & Edge F1 $\uparrow$ & 0.165 & 0.200 & 0.403 & 0.341 & \textbf{0.510} \\
\midrule
\multicolumn{7}{l}{\emph{Additional diagnostics}} \\
\midrule
3DHarnessBench & Uni3D $\uparrow$ & -- & 0.257 & 0.266 & -- & \textbf{0.649} \\
\midrule
CC-HARD & Text $\uparrow$ & -- & 0.831 & 0.926 & -- & \textbf{0.974} \\
 & Position $\uparrow$ & -- & 0.826 & 0.868 & -- & \textbf{0.935} \\
 & Visual CLIP $\uparrow$ & -- & 0.849 & 0.849 & -- & \textbf{0.888} \\
 & MAE $\downarrow$ & -- & 42.957 & 35.773 & -- & \textbf{21.749} \\
\midrule
VFIG-Bench (ID) & DINOv2-base $\uparrow$ & -- & 0.960 & 0.963 & -- & \textbf{0.989} \\
 & CLIP-B/32 $\uparrow$ & -- & 0.952 & 0.955 & -- & \textbf{0.981} \\
 & SigLIP2 $\uparrow$ & -- & 0.983 & 0.984 & -- & \textbf{0.993} \\
\midrule
DrawAI-Bench & Image source recall $\uparrow$ & -- & 0.022 & 0.035 & -- & \textbf{0.065} \\
 & Image IoU $\uparrow$ & -- & \textbf{0.000} & \textbf{0.000} & -- & \textbf{0.000} \\
\bottomrule
\end{tabular*}
\end{ArxivInlineBlock}

\paragraph{Agentic RL gains over SFT.}
Starting from the supervised policy, agentic RL rewards programs that execute and reproduce the target observations. Execution success on Code2Video-Compass rises from 0.472 to 0.979, alongside gains in appearance fidelity (0.666 to 0.837) and motion fidelity (0.458 to 0.520). On VFIG-Bench, render success more than doubles, from 0.488 to 1.000. Learning from execution makes the gains visible in both playable videos and executable vector graphics.

\paragraph{What the RL-trained policy retains in a single pass.}
CodingDINO-Single and CodingDINO share the same weights. The first writes the program directly from the input; the second can inspect renders and revise. Across the 12 headline metrics, the first draft retains 85.4\% of multi-turn quality on average and exceeds the teacher on ten of the twelve metrics. On 3DHarnessBench, the first draft already matches the target across viewpoints (0.866 SigLIP2 vs.\ 0.887 after revision). On VFIG-Bench it renders successfully in 99.8\% of cases. Where individual elements move or contours must align precisely, feedback adds a further gain: AnimWeb-Bench velocity improves from 0.241 to 0.371 and DrawAI-Bench Edge F1 from 0.341 to 0.510. Learning from rendered outcomes thus improves the program the policy writes before it sees any render at inference.

Table~\ref{tab:supp-extra-checkpoint-splits} compares Base and SFT on valid 3D meshes and on out-of-distribution and pooled vector graphics. On out-of-distribution SVGs, supervision raises SSIM from 0.688 to 0.783.

\begin{ArxivInlineBlock}
\captionsetup{type=table}
\centering
\caption{Additional checkpoint comparisons on valid 3DHarnessBench outputs and the out-of-distribution (OOD) and pooled VFIG-Bench splits. The 3D panel reports mean geometric errors over valid outputs.}
\label{tab:supp-extra-checkpoint-splits}
\footnotesize
\setlength{\tabcolsep}{3pt}
\begin{tabular*}{\textwidth}{@{\extracolsep{\fill}}lrrrrrrr@{}}
\toprule
\rowcolor{cdbench}
\multicolumn{8}{c}{\textbf{3DHarnessBench (valid outputs)}} \\
\midrule
Model & SigLIP2 $\uparrow$ & DINOv2-L $\uparrow$ & DINOv3-L $\uparrow$ & Uni3D $\uparrow$ & Uni3D I-3D $\uparrow$ & \shortstack{Chamfer\\mean} $\downarrow$ & \shortstack{Betti L1\\mean} $\downarrow$ \\
\midrule
CodingDINO-Base & \textbf{0.834} & \textbf{0.492} & \textbf{0.565} & 0.476 & \textbf{0.272} & \textbf{0.035} & 646.751 \\
CodingDINO-SFT & 0.831 & 0.452 & 0.548 & \textbf{0.492} & 0.271 & 0.041 & \textbf{12.707} \\
\bottomrule
\end{tabular*}\par
\begin{tabular*}{\textwidth}{@{\extracolsep{\fill}}lrrrrrrrr@{}}
\toprule
\rowcolor{cdbench}
\multicolumn{9}{c}{\textbf{VFIG-Bench (OOD)}} \\
\midrule
Model & SSIM $\uparrow$ & 1/LPIPS $\uparrow$ & DINOv2-base $\uparrow$ & CLIP-B/32 $\uparrow$ & SigLIP2 $\uparrow$ & VisualSim $\uparrow$ & Clean $\uparrow$ & Render $\uparrow$ \\
\midrule
CodingDINO-Base & 0.688 & 2.974 & 0.902 & 0.913 & 0.965 & 0.926 & \textbf{0.800} & \textbf{0.990} \\
CodingDINO-SFT & \textbf{0.783} & \textbf{4.913} & \textbf{0.926} & \textbf{0.942} & \textbf{0.976} & \textbf{0.948} & 0.655 & 0.374 \\
\bottomrule
\end{tabular*}\par
\begin{tabular*}{\textwidth}{@{\extracolsep{\fill}}lrrrrrrrr@{}}
\toprule
\rowcolor{cdbench}
\multicolumn{9}{c}{\textbf{VFIG-Bench (pooled)}} \\
\midrule
Model & SSIM $\uparrow$ & 1/LPIPS $\uparrow$ & DINOv2-base $\uparrow$ & CLIP-B/32 $\uparrow$ & SigLIP2 $\uparrow$ & VisualSim $\uparrow$ & Clean $\uparrow$ & Render $\uparrow$ \\
\midrule
CodingDINO-Base & 0.783 & 4.178 & 0.941 & 0.939 & 0.977 & 0.952 & \textbf{0.790} & \textbf{0.988} \\
CodingDINO-SFT & \textbf{0.839} & \textbf{6.458} & \textbf{0.953} & \textbf{0.952} & \textbf{0.981} & \textbf{0.962} & 0.701 & 0.450 \\
\bottomrule
\end{tabular*}\par
\end{ArxivInlineBlock}

\Needspace{4\baselineskip}
\subsection{What depth contributes to the reward}
\label{sec:supp-depth}
Adding relative depth to the two-encoder reward reduces mean Betti error by 45.6\% on 3DHarnessBench, from 12.930 to 7.034. Multi-view appearance also improves, with SigLIP2 rising from 0.879 to 0.882 (Table~\ref{tab:reward}). On AnimWeb-Bench the same training change improves motion position, size, velocity and CLIP. On DrawAI-Bench, text-box recall rises from 0.416 to 0.483 and formula editability rises from 0.723 to 0.873.
\section{Agentic training and evaluation settings}
\label{sec:supp-training}

\Needspace{4\baselineskip}
\subsection{Prompts for visual reconstruction}
\label{sec:supp-harness}
The shared system prompt directs the policy to inspect motion across frames and prioritize the largest discrepancy after each render. Task prompts focus revision on features specific to each representation, including part geometry in Blender and text spacing in SVG.

\begin{tcolorbox}[colback=templatebackground,colframe=templatebackground,boxrule=0pt,arc=3pt,left=8pt,right=8pt,top=6pt,bottom=6pt]
\small\raggedright
\textbf{System prompt: temporal inspection}\par
\emph{For video, compare the beginning, middle and end and inspect neighboring frames around motion changes. Identify object trajectories, relative positions, timing, transitions, occlusions and any camera movement.}\smallskip

\textbf{System prompt: feedback-guided revision}\par
\emph{Use these observations to correct the largest visible discrepancy, preserve matching parts, then render the updated code and verify the affected regions and motion again.}\smallskip

\textbf{Blender task prompt: geometry}\par
\emph{After rendering, compare the projected outline and the gaps between parts first. Check the width-to-height ratio, bending direction, thickness, endpoints and contacts.}\smallskip

\textbf{SVG task prompt: typography}\par
\emph{Preserve text scale, style and line spacing relative to surrounding shapes.}
\end{tcolorbox}

\Needspace{4\baselineskip}
\subsection{Supervision and policy optimization}
\label{sec:supp-policy-sft}
The 27B policy has a vision encoder of width 1,152 and a 64-layer language backbone of width 5,120 that combines gated delta-rule layers with causal attention. During supervised fine-tuning, the loss mask selects the teacher's actions across a verified trajectory, including edits made after an unsuccessful attempt. Target images, intermediate observations and execution diagnostics condition the next action. A single pretrained multi-token prediction layer supplies an auxiliary loss at three prediction depths, with a combined weight of 0.1.

\begin{ArxivInlineBlock}
\captionsetup{type=table}
\centering\small
\caption{Training settings at two GPU allocations. Global batch is measured in trajectories per optimizer update. GRPO entries give targets $\times$ rollouts per target.}
\label{tab:supp-optimization}
\setlength{\tabcolsep}{3pt}
\begin{tabular*}{0.86\linewidth}{@{\extracolsep{\fill}}lcc@{}}
\toprule
\rowcolor{cdbench}
Setting & SFT & Multi-turn GRPO \\
\midrule
\multicolumn{3}{l}{\emph{Global batch by GPU allocation}} \\
512 GPUs ($4\times$) & 256 & $128\times8=1{,}024$ \\
1,536 GPUs ($8\times$) & 512 & $256\times8=2{,}048$ \\
\midrule
Context (tokens) & 204,800 & 131,072 \\
Initial learning rate & $10^{-6}$ & $5\times10^{-7}$ \\
Peak learning rate & $10^{-5}$ & $10^{-6}$ \\
After warmup & Cosine to $10^{-6}$ & Constant \\
TP / PP / CP & $4/2/2$ & $4/1/4$ \\
\midrule
Microbatch / precision & \multicolumn{2}{c}{1 trajectory / BF16} \\
Optimizer & \multicolumn{2}{c}{Distributed Adam} \\
$(\beta_1,\beta_2)$ / $\epsilon$ & \multicolumn{2}{c}{$(0.9,0.999)$ / $10^{-8}$} \\
Weight decay / grad.\ clip & \multicolumn{2}{c}{0.01 / 1.0} \\
\bottomrule
\end{tabular*}
\end{ArxivInlineBlock}

Table~\ref{tab:supp-optimization} gives the learner settings. TP, PP and CP denote tensor, pipeline and context parallelism. The GRPO learning rate reaches its peak after 13 updates. The batch multipliers use 64 trajectories for SFT and 256 for GRPO as their reference. With the depth-inclusive reward, GRPO~\cite{shao2024deepseekmath} compares the eight rollouts for each target. Failed episodes enter that comparison with zero reward. We add $10^{-6}$ to the group's sample standard deviation when scaling the leave-one-out advantage. The token-level policy update clips the current-to-previous policy ratio to $[0.8,1.2]$ and uses a dual-clip coefficient of 3 for negative advantages. A separate importance weight corrects for the difference between rollout and learner probabilities, with a cap of 2. The reference-policy KL coefficient is zero.

\begin{ArxivInlineBlock}
\captionsetup{type=table}
\centering\small
\caption{Online rollout sampling and held-out validation.}
\label{tab:supp-rollout-settings}
\begin{tabularx}{0.86\linewidth}{@{}Xrr@{}}
\toprule
\rowcolor{cdbench}
Setting & Training & Validation \\
\midrule
Rollouts per target & 8 & 1 \\
Sampling temperature & 1.0 & 0.6 \\
\midrule
Nucleus sampling $p$ & \multicolumn{2}{c}{0.95} \\
Explicit thinking mode & \multicolumn{2}{c}{Disabled} \\
Output tokens per turn (cap) & \multicolumn{2}{c}{16,384} \\
Renders per episode (cap) & \multicolumn{2}{c}{6} \\
Policy turns / tool calls (caps) & \multicolumn{2}{c}{100 / 100} \\
Video sampling / frame cap & \multicolumn{2}{c}{2 fps / 128 frames} \\
Video input long edge (cap) & \multicolumn{2}{c}{768 pixels} \\
\bottomrule
\end{tabularx}
\end{ArxivInlineBlock}

Online rollouts use four-way tensor parallelism and three-token speculative decoding. The asynchronous sampler accepts trajectories at most one policy update behind the learner. The depth-inclusive recipe allows up to 100 optimizer updates. We evaluate the same 128 targets before training and every five updates, using the sampling settings in Table~\ref{tab:supp-rollout-settings}. Each validation point averages terminal rewards across all 128 rollouts. The training curve also shows an exponential moving average with weight 0.7 on the previous smoothed value.

\clearpage
\subsection{Appearance and relative-depth rewards}
\label{sec:supp-visual-consistency}
For each target--render pair, frozen DINOv2-L~\cite{oquab2023dinov2} and DINOv3-L/16~\cite{simeoni2025dinov3} encode the complete frame at $448\times448$ pixels. The appearance scores $a_2$ and $a_3$ are cosine similarities between the corresponding final-layer CLS tokens. For Three.js and Blender, we also use Depth Anything V2 Small to estimate relative depth in each image. We center each map at its own median and divide by its mean absolute deviation from that median plus $10^{-6}$. For normalized target and rendered depth maps $Z,\widehat Z$ on a pixel grid $\Omega$, the depth agreement is
\begin{equation}
 d=\operatorname{clip}_{[0,1]}\!\left(
  1-\frac{1}{2|\Omega|}\sum_{x\in\Omega}|Z(x)-\widehat Z(x)|
 \right).
\end{equation}
Independent normalization makes foreground--background separation comparable across the two predictions. Three.js and Blender receive the per-frame score $(a_2+a_3+d)/3$. Other program types receive $(a_2+a_3)/2$. For videos, the terminal reward averages these scores at matched timestamps, giving each trajectory one training signal regardless of clip length.

\Needspace{4\baselineskip}
\subsection{Inference measurements}
\label{sec:supp-inference-measurements}
The quality panel uses 70 paired cases: 50 selected for matching the teacher within 0.01 on every recorded metric with fewer renders and tokens, plus 20 additional lower-token cases. At each render budget we count a completed program if its largest per-metric drop from the teacher's final score, accounting for metric direction, is at most 0.01. The shaded band in Figure~\ref{fig:budget} extends the tolerance to 0.03. The teacher curve counts final programs completed by each render budget.

\begin{ArxivInlineBlock}
\captionsetup{type=table}
\centering\small
\caption{Quality and token use across render budgets. Quality counts completed programs meeting the teacher's final quality within the indicated tolerance. Cost is mean cumulative input and output tokens in millions.}
\label{tab:supp-budget-cost}
\begin{tabular*}{0.86\linewidth}{@{\extracolsep{\fill}}lrrrr@{}}
\toprule
\rowcolor{cdbench}
Render budget & 1 & 2 & 4 & 6 \\
\midrule
\multicolumn{5}{l}{\emph{Programs reaching teacher quality (70 selected pairs)}} \\
CodingDINO, tolerance $0.01$ & 1 & 3 & 32 & 52 \\
CodingDINO, tolerance $0.03$ & 1 & 8 & 38 & 62 \\
Teacher, final program & 1 & 3 & 6 & 70 \\
\midrule
\multicolumn{5}{l}{\emph{Mean tokens in millions (90 fixed pairs)}} \\
CodingDINO & 0.110 & 0.241 & 0.528 & 0.666 \\
Teacher & 0.757 & 1.392 & 2.635 & 3.389 \\
\bottomrule
\end{tabular*}
\end{ArxivInlineBlock}

The token panel uses a fixed set of 90 paired tasks and includes repeated input context and generated reasoning tokens. At four renders, 32 CodingDINO programs meet the stricter quality criterion, while six teacher runs have finished. On the 90 paired cost cases, CodingDINO uses one fifth of the teacher's mean token count at the same budget (0.528M vs.\ 2.635M, Table~\ref{tab:supp-budget-cost}).
\FloatBarrier
\clearpage
\section{Qualitative analysis}
\label{sec:supp-qualitative}
Each row shows the input followed by renders from CodingDINO, its teacher DeepSeek-V4.1-Flash (\texttt{high} thinking effort), and its base model Qwen3.8-27B (\texttt{medium} thinking effort).

\paragraph{Viewpoint in Three.js scenes.}
The rowing scene in Figure~\ref{fig:qual-threejs} shows how viewpoint changes the spatial relations a program must encode. In the input, the boat runs from lower left to upper right, with oars on both sides. Qwen shows the boat almost broadside. DeepSeek reconstructs an oblique view but reverses the diagonal. CodingDINO preserves both the diagonal and the arrangement of the oars. The same attention to viewpoint is visible in the motorhome's rear and side windows.

\begin{ArxivInlineBlock}
\captionsetup{type=figure}
\centering
\includegraphics[width=\linewidth,height=0.66\textheight,keepaspectratio]{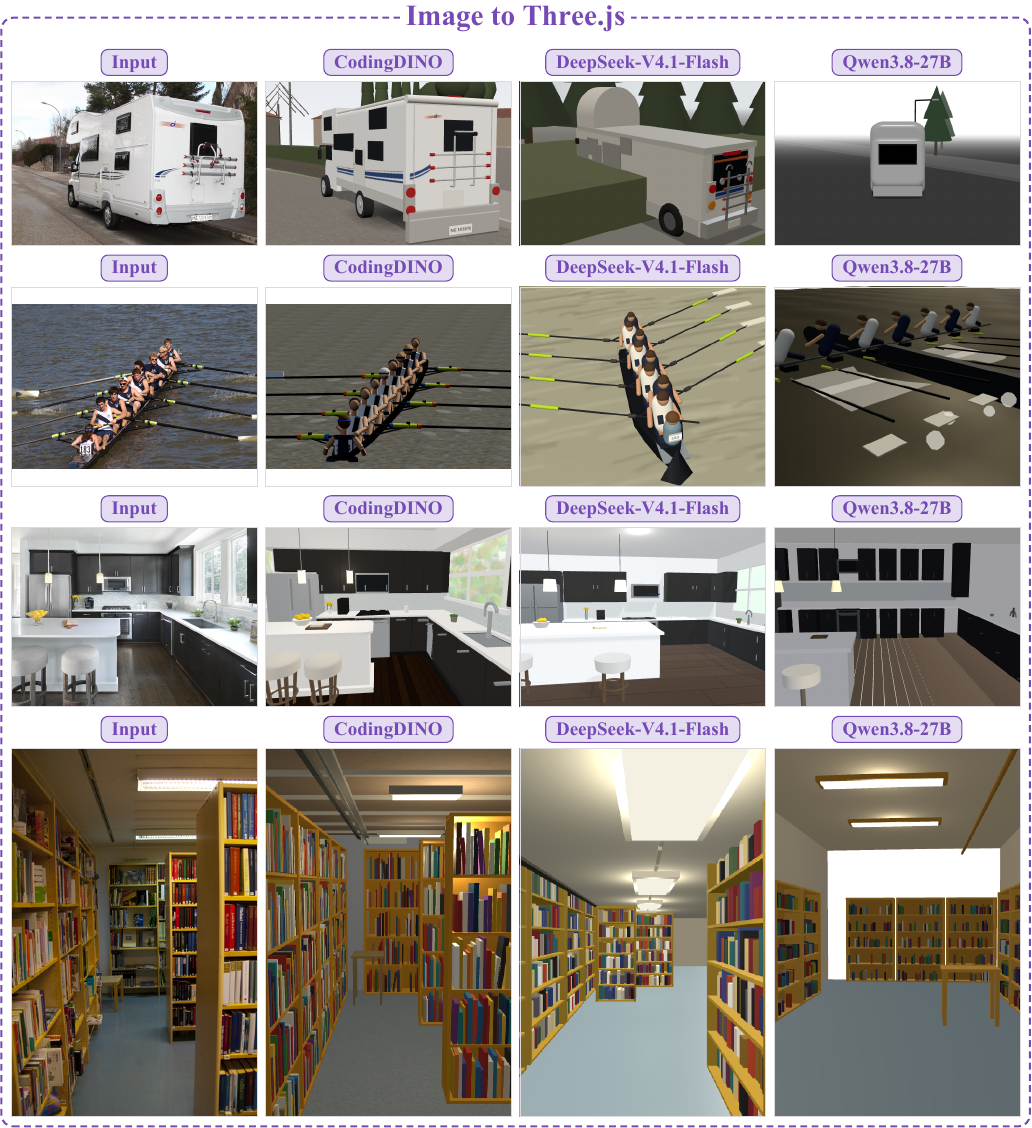}
\caption{\textbf{Viewpoint and layout in Three.js.} From top to bottom: a motorhome, a rowing crew, a kitchen and a library aisle. The motorhome windows and oars reveal the viewing angle; the countertops and shelving reveal how the scene recedes from the camera.}
\label{fig:qual-threejs}
\end{ArxivInlineBlock}

\clearpage
\paragraph{Part arrangement in Blender objects.}
The hook assembly in Figure~\ref{fig:qual-blender} tests the placement and orientation of repeated parts. Qwen places the purple support on the right. DeepSeek puts it on the left, as in the input, but its hooks curve upward. CodingDINO recovers the left support and three downward-hanging hooks beneath the panel. The cabinet shows a similar correspondence between parts: open shelves sit above the left end of a continuous white countertop.

\begin{ArxivInlineBlock}
\captionsetup{type=figure}
\centering
\includegraphics[width=\linewidth,height=0.76\textheight,keepaspectratio]{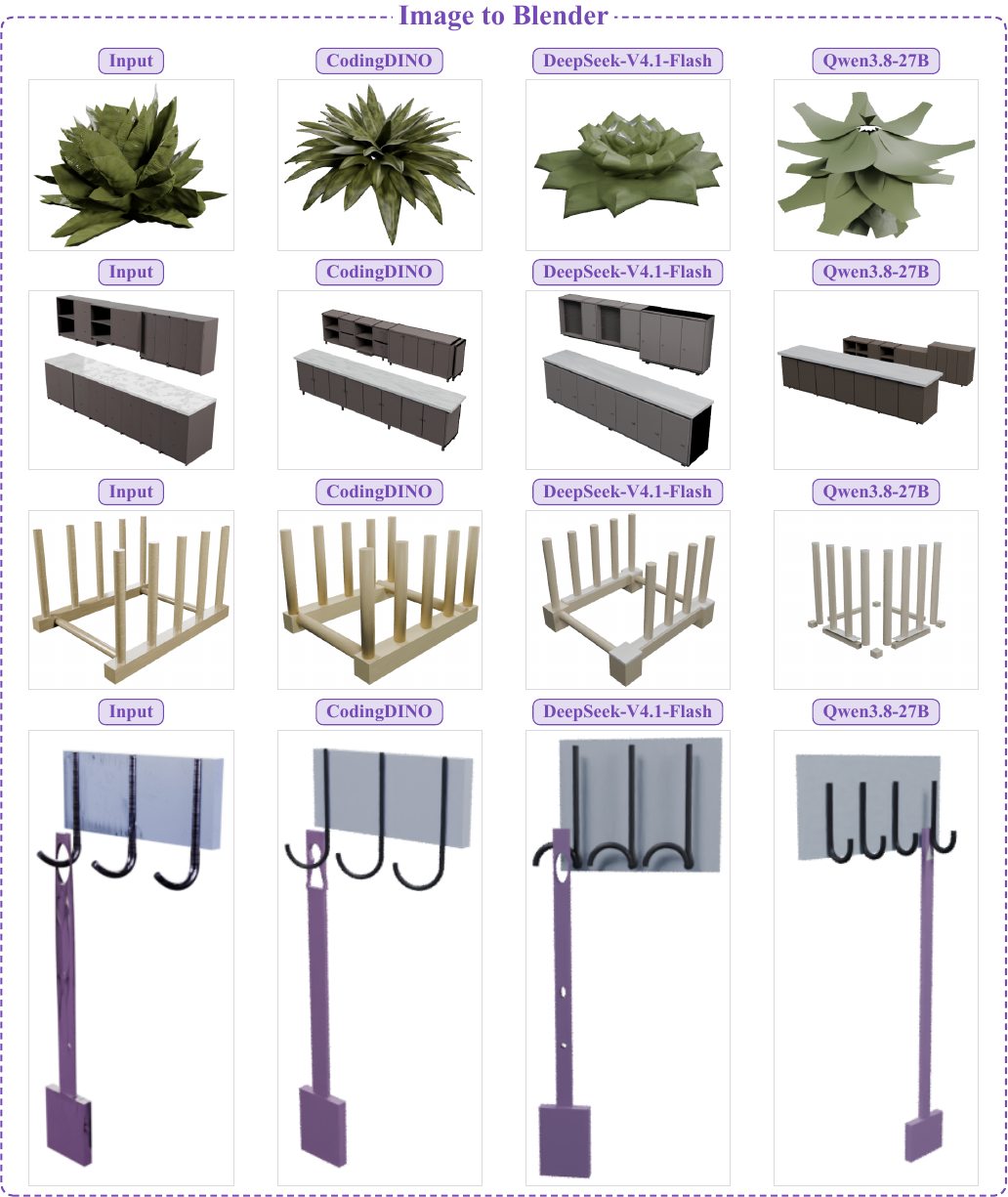}
\caption{\textbf{Object structure in Blender.} From top to bottom: a leafy plant, a cabinet set, a wooden frame with upright posts and a hook assembly. The open compartments, repeated posts and hanging hooks make the placement and orientation of individual parts visible.}
\label{fig:qual-blender}
\end{ArxivInlineBlock}

\clearpage
\paragraph{Page layout and diagram geometry.}
On the HTML article page in Figure~\ref{fig:qual-html-svg}, Qwen places solid color blocks beside the article, while CodingDINO recovers a column of distinct thumbnails and the speaker's outstretched hand. In the two-branch SVG, DeepSeek follows the branches but draws the yellow blocks too narrowly. CodingDINO preserves their width relative to the arrows and the different sizes of the blue outputs. These examples show what the advantage over the base and teacher can look like in a rendering: the program preserves orientation, attachment and relative scale as well as the identity of the objects.

\begin{ArxivInlineBlock}
\captionsetup{type=figure}
\centering
\includegraphics[width=\linewidth,height=0.76\textheight,keepaspectratio]{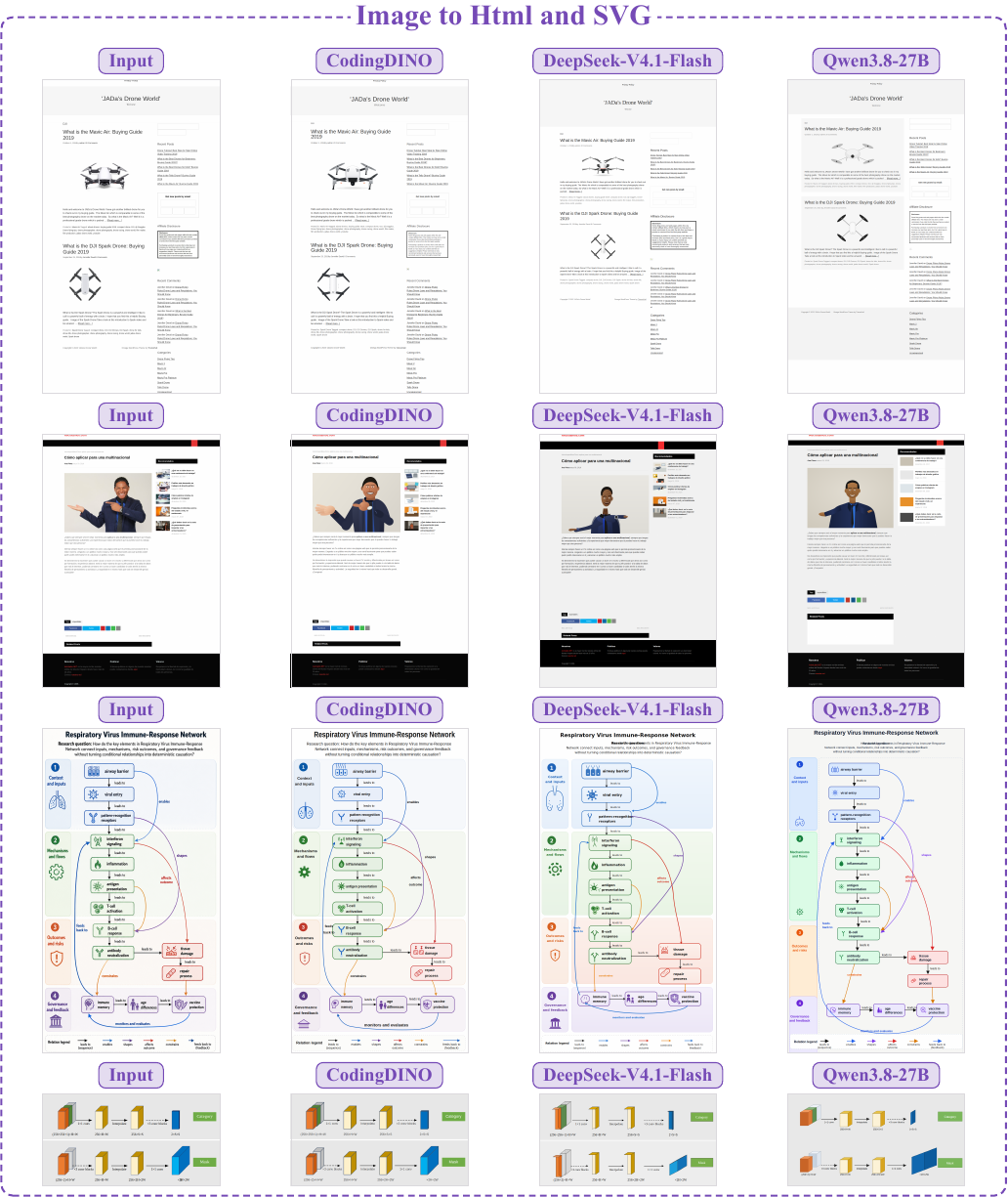}
\caption{\textbf{Web pages in HTML and diagrams in SVG.} The first two rows show a drone website and an article page. The last two show a biological pathway diagram and a two-branch network diagram. The sidebar thumbnails and the widths of connected diagram blocks provide local references for scale and alignment.}
\label{fig:qual-html-svg}
\end{ArxivInlineBlock}
\clearpage
\section{Toward an editable world model}
\label{sec:editable-world-model}
\emph{Must we train a video generator to build one?} \emph{What if a world model could emerge inside an LLM?}

\noindent\textbf{Let the LLM edit the world it wrote.} Raise the camera over the freeway, widen the carton in the robot's arms, or leave a single horse in the field. Each edit changes one line in the Three.js program CodingDINO wrote from the input video. Figure~\ref{fig:world-model-edits} shows the same instant before and after the edit.

\smallskip
\begin{ArxivInlineBlock}
\captionsetup{type=figure}
\centering
\includegraphics[width=\linewidth]{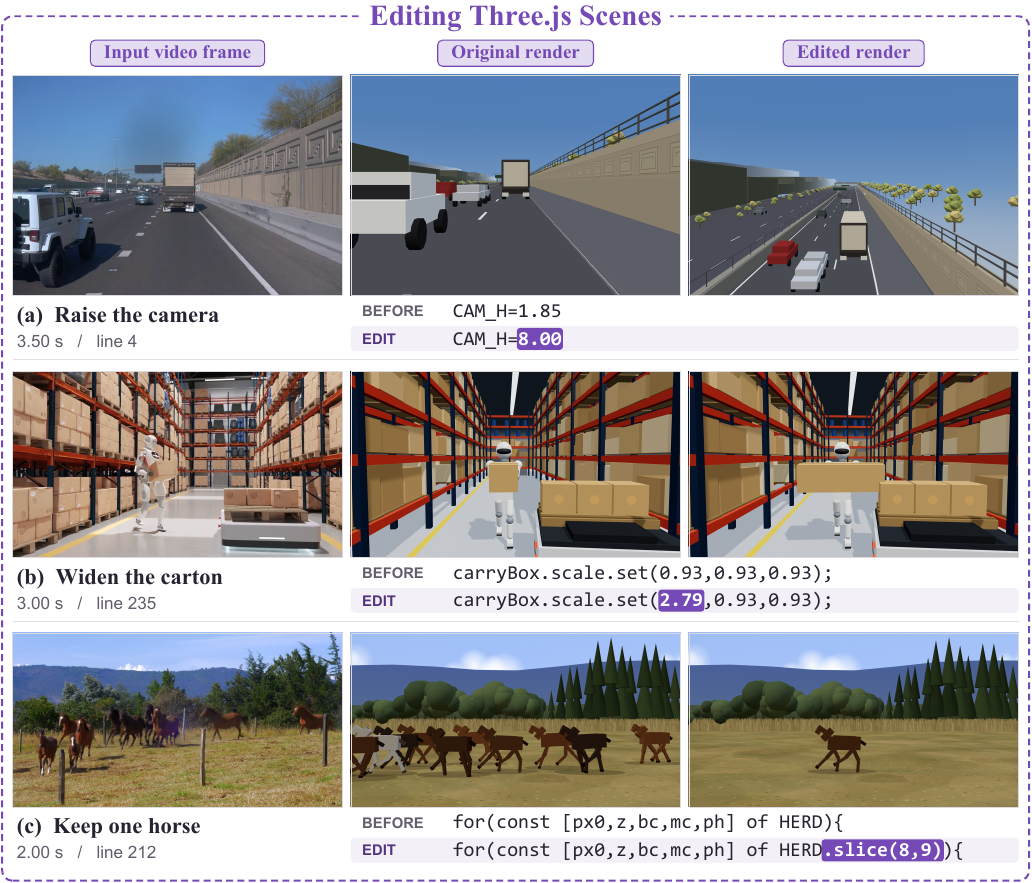}
\caption{\textbf{Editing the world written by CodingDINO.} Each row pairs an input frame with the original and edited renders at the indicated time. The code excerpts identify the change to the camera, the carton, or the herd.}
\label{fig:world-model-edits}
\end{ArxivInlineBlock}
\smallskip

Raising the freeway camera brings the truck's roof and more of the road ahead into view. The camera still faces the same direction, and the vehicles follow their original trajectories. In the warehouse, tripling the carton's width along its local $x$-axis makes it extend beyond both sides of the robot's torso as the robot carries it along the same path. In the field, selecting the ninth of thirteen horses removes its twelve neighbors. The remaining horse follows its original trajectory and, at the matched instant, is in the same phase of its stride against the same tree line.

These edits suggest a route to a world model inside an LLM. Reconstructing a video asks the model to account for what it saw. Editing the reconstruction asks which relationships should survive an intervention. The LLM can express its answer in code, run the scene forward under the new condition, and inspect the outcome.

\addtocontents{toc}{\protect\setcounter{tocdepth}{2}}

\end{document}